\pdfoutput=1

\documentclass[11pt]{article}

\usepackage{acl}

\usepackage{times}
\usepackage{latexsym}

\usepackage[T1]{fontenc}

\usepackage[utf8]{inputenc}

\usepackage{microtype}

\usepackage{amsmath,amsfonts,bm}

\def\eqref#1{equation~\ref{#1}}

\def\1{\bm{1}}

\DeclareMathAlphabet{\mathsfit}{\encodingdefault}{\sfdefault}{m}{sl}
\SetMathAlphabet{\mathsfit}{bold}{\encodingdefault}{\sfdefault}{bx}{n}

\definecolor{citecolor}{RGB}{34,139,34} 
\definecolor{linkcolor}{HTML}{ED1C24}

\usepackage{pifont}

\usepackage{graphicx}               
\usepackage{tabularx}               
\newcolumntype{C}{>{\centering\arraybackslash}X}
\usepackage{multirow}               
\usepackage{diagbox}                
\usepackage{hhline}                 
\usepackage{color}                  
\usepackage{xcolor}                  
\usepackage{amsmath}                
\usepackage{amsfonts}                
\usepackage{amssymb}                
\usepackage{mathtools}              
\usepackage{enumitem}               
\usepackage[ruled,vlined]{algorithm2e}
\usepackage{tabu}            
\usepackage{booktabs} 
\usepackage{array}  
\usepackage{caption}  
\usepackage{subcaption} 

\newtheorem{thm:def}{Definition}
\newtheorem{thm:eg}{Example}
\newtheorem{thm:lem}{Lemma}

\definecolor{RoseQuartzBg}{HTML}{F7CAC9}
\definecolor{RoseQuartz}{HTML}{F5A798}
\definecolor{Serenity}{HTML}{92A8D1}
\definecolor{OrangeRed}{rgb}{1.0, 0.27, 0.0}
\definecolor{Turquoise}{HTML}{0F4C81}
\definecolor{mint}{rgb}{0.24, 0.71, 0.54}
\definecolor{byzantine}{rgb}{0.74, 0.2, 0.64}
\definecolor{byzantium}{rgb}{0.44, 0.16, 0.39}
\definecolor{captioningtask}{HTML}{9C843F}
\definecolor{qatask}{HTML}{CC6600}
\definecolor{temporalmarker}{HTML}{7F00FF}
\definecolor{targettext}{HTML}{3333FF}
\definecolor{prompttext}{HTML}{666666}
\definecolor{videolevel}{HTML}{330066}
\definecolor{framelevel}{HTML}{0066CC}
\definecolor{tokenlevel}{HTML}{336600}
\definecolor{boxgrey}{HTML}{666666}
\definecolor{boxblue}{HTML}{6C8EBF}
\definecolor{boxgreen}{HTML}{82B366}
\definecolor{textgreen}{HTML}{009900}
\definecolor{textred}{HTML}{FF0000}
\definecolor{textreddark}{HTML}{CC0000}
\definecolor{textblue}{HTML}{0066CC}
\definecolor{cmtwzhl}{HTML}{8E44AD}
 
\usepackage{wrapfig}
\usepackage{xparse}

\newcommand{\addcheckemoji}{\raisebox{-.2\height}{\includegraphics[height=12pt]{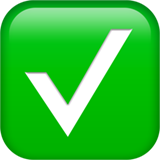}}}
\newcommand{\addcrossemoji}{\raisebox{-.2\height}{\includegraphics[height=12pt]{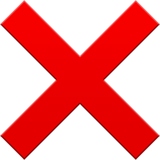}}}

\NewDocumentCommand{\zhenhailong}
{ mO{} }{\textcolor{cmtwzhl}{\textsuperscript{\textit{Zhenhailong}}\textsf{\textbf{\small[#1]}}}}

\NewDocumentCommand{\heng}
{ mO{} }{\textcolor{red}{\textsuperscript{\textit{Heng}}\textsf{\textbf{\small[#1]}}}}

\newcommand{\ours}{Solo Performance Prompting} 
\newcommand{\oursabbr}{\text{SPP}}
\newcommand{\oursproabbr}{\text{SPP-Profile}}
\newcommand{\oursfixabbr}{\text{SPP-Fixed-Persona}}
\newcommand{\tasktrivia}{Trivia Creative Writing}
\newcommand{\taskcodenames}{Codenames Collaborative}
\newcommand{\tasklogic}{Logic Grid Puzzle}

\usepackage{adjustbox} 

\title{Unleashing the Emergent Cognitive Synergy in Large Language Models: \\ A Task-Solving Agent through Multi-Persona Self-Collaboration}

\author{%
Zhenhailong Wang\textsuperscript{\textnormal{1}\thanks{\;\;Work was done when interning at Microsoft Research Asia.}},
Shaoguang Mao\textsuperscript{\textnormal{2}}, 
Wenshan Wu\textsuperscript{\textnormal{2}}, 
Tao Ge\textsuperscript{\textnormal{2}\thanks{\;\;Corresponding author.}},
Furu Wei\textsuperscript{\textnormal{2}}, 
Heng Ji\textsuperscript{\textnormal{1}}\\ 
\textsuperscript{1}University of Illinois Urbana-Champaign, \textsuperscript{2}Microsoft Research Asia\\
\texttt{\{wangz3,hengji\}@illinois.edu}\\ \texttt{\{shaoguang.mao,wenshan.wu,tage,fuwei\}@microsoft.com}
}

\begin{document}

\maketitle

\begin{abstract}
Human intelligence thrives on cognitive synergy, where collaboration among different minds yield superior outcomes compared to isolated individuals.
In this work, we propose \textbf{\ours{} (\oursabbr{})}, which transforms a single LLM into a \textbf{cognitive synergist} by engaging in multi-turn self-collaboration with multiple personas. 
%
A cognitive synergist is an intelligent agent that collaboratively combines multiple minds' strengths and knowledge to enhance problem-solving in complex tasks.
By dynamically identifying and simulating different personas based on task inputs, \oursabbr{} unleashes the potential of cognitive synergy in LLMs. 
Our in-depth analysis shows that assigning multiple fine-grained personas in LLMs improves problem-solving abilities compared to using a single or fixed number of personas.
We evaluate \oursabbr{} on three challenging tasks: Trivia Creative Writing, Codenames Collaborative, and Logic Grid Puzzle, encompassing both \textbf{knowledge-intensive} and \textbf{reasoning-intensive} types. Unlike previous works, such as Chain-of-Thought, that solely enhance the reasoning abilities in LLMs, experimental results demonstrate that \oursabbr{} effectively reduces factual hallucination, and maintains strong reasoning capabilities. Additionally, comparative experiments show that cognitive synergy only \textbf{emerges} in GPT-4 and does not appear in less capable models, such as GPT-3.5-turbo and Llama2-13b-chat, which draws an interesting analogy to human development.
%
Code, data, and prompts can be found at: \url{https://github.com/MikeWangWZHL/Solo-Performance-Prompting.git}
\end{abstract}

\begin{figure}[ht]
  \centering
  \includegraphics[width=0.45\textwidth]{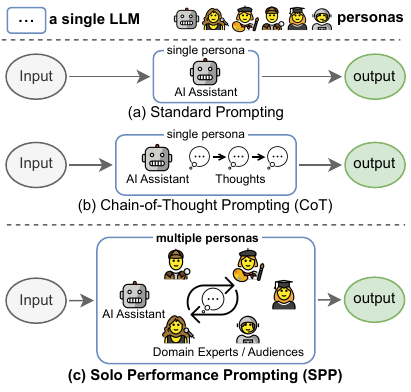}
  \caption{
    Schematic illustration of \ours{} (\oursabbr{}) and the difference compared to previous prompting methods. 
    \vspace{-10pt}
  }
  \label{fig:teaser_figure}
\end{figure}

\section{Introduction} 

\begin{figure*}[th]
  \centering
  \includegraphics[width=0.92\textwidth]{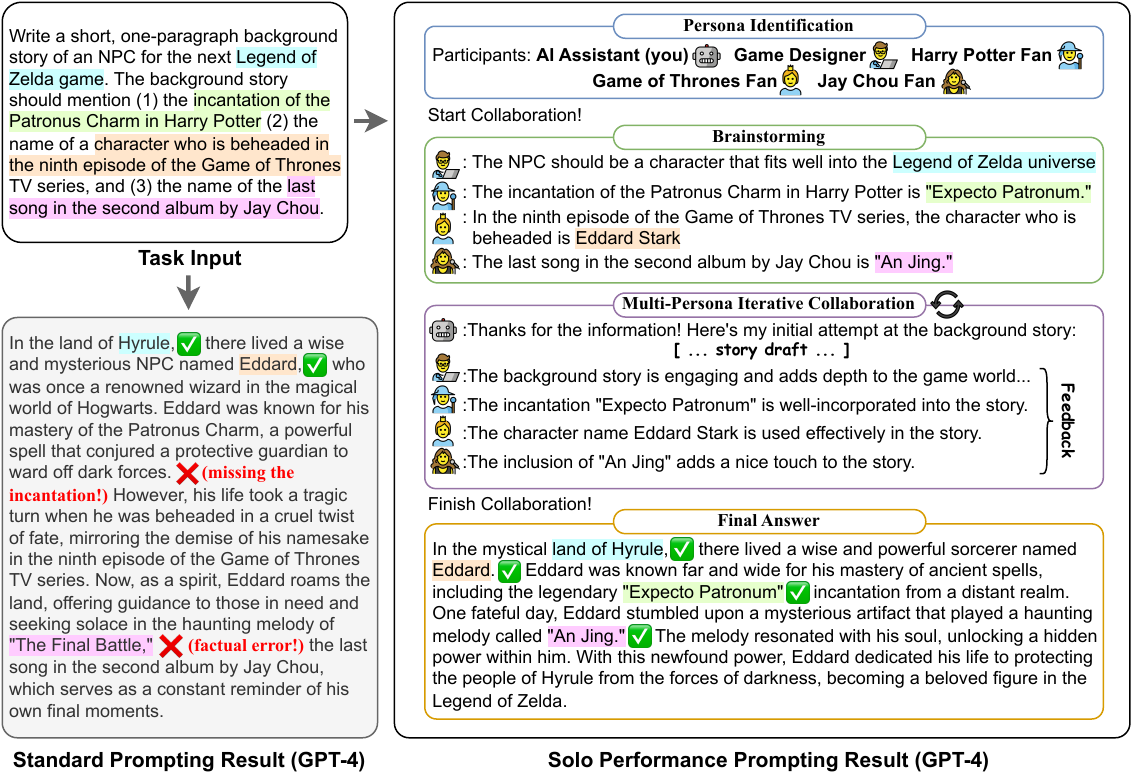}
  \caption{
  Task-solving example of \ours{} (\oursabbr{}) with GPT-4. The personas of the participants are automatically identified by GPT-4 based on the task input. This example shows that Standard Prompting suffers from factual errors, whereas \oursabbr{} provides accurate information and a coherent answer. Note that, in real-world applications, the domains can vary not only within entertainment but also encompass history, science, education, healthcare, etc.
  \vspace{-10pt}
  }
  \label{fig:main_illustration}
\end{figure*}

Although large language models (LLMs) have demonstrated impressive performance as general task-solving agents, they still encounter challenges~\citep{qin2023chatgpt, chatgpt-analysis, openai2023gpt4, gpt4-early-exp} in various knowledge-intensive and reasoning-intensive tasks due to factual hallucination~\citep{maynez-etal-2020-faithfulness} and a lack of slow-thinking~\citep{sloman1996empirical} capabilities.
Unlike humans, who can leverage the power of collaboration and information integration among different cognitive processes and individuals (referred to as \textit{cognitive synergy}~\citep{curcseu2015cognitive, goertzel2009cognitive, goertzel2017formal}), current LLMs are akin to "jack-of-all-trades" with a vast mixture of knowledge and characteristics. 
Recent advancements, such as Chain-of-Thought (CoT) prompting~\citep{wei2023chainofthought, think-step-by-step} and Self-refinement~\citep{madaan2023self_refine, shinn2023reflexion}, have successfully enhanced the reasoning abilities of LLMs by simulating slow-thinking through the generation of intermediate steps or iterative revision. However, factual hallucination remains a major challenge for LLMs on knowledge-intensive tasks.


A cognitive synergist is an intelligent agent that collaborates with multiple minds to enhance problem-solving and efficacy in complex tasks.
In this work, we aim to \textbf{create a cognitive synergist based on a single LLM} that can \textit{"split into" multiple personas and engage in self-collaboration to solve both knowledge-intensive and reasoning-intensive tasks}. 
This idea is heavily inspired by the role of pretend play~\citep{piaget1954construction,role_of_play2009} in cognitive development and recent findings that assigning personas~\citep{deshpande2023toxicity, xu2023expertprompting} to LLMs can elicit specific behaviors, improve answer quality, and potentially build an AI society~\cite{generative_agents, schick2022peer, li2023camel, toolmakers} with collaborative LLM agents. However, as shown in Table~\ref{tab:comparison}, previous works have limitations such as fixed or task-specific personas, the need for additional fine-tuning, and increased inference costs due to multiple LLM instances.

\begin{table*}[th!]
\centering
\resizebox{0.95\textwidth}{!}{
\begin{tabular}{@{}lcccccc@{}}
    \toprule 
    &   \textbf{\begin{tabular}[c]{@{}c@{}}General task \\ solving? \end{tabular}} &
        \textbf{\begin{tabular}[c]{@{}c@{}}Pure zero-shot \\ prompting? \end{tabular}} &
        \textbf{\begin{tabular}[c]{@{}c@{}}Has multiple \\ personas? \end{tabular}} &
        \textbf{\begin{tabular}[c]{@{}c@{}}Personas dynamically \\ identified?\end{tabular}} &
        \textbf{\begin{tabular}[c]{@{}c@{}}Has iterative \\ refinement?\end{tabular}} &
        \textbf{\begin{tabular}[c]{@{}c@{}}Need only a \\ single LLM?\end{tabular}} \\
      
    \midrule
        $\dagger$ Standard Prompting~\cite{gpt3} & \addcheckemoji{} & \addcheckemoji{} & \addcrossemoji{}              & \addcrossemoji{} & \addcrossemoji{} & \addcheckemoji{}  \vspace{1pt}\\
        $\dagger$  Chain-of-Thought~\citep{wei2023chainofthought} & \addcheckemoji{} & \addcheckemoji{} & \addcrossemoji{}              & \addcrossemoji{} & \addcrossemoji{} & \addcheckemoji{}  \vspace{1pt}\\
        Inner Monologue~\citep{inner-monologue}         & \addcrossemoji{} & \addcheckemoji{} & \addcrossemoji{}              & \addcrossemoji{} & \addcheckemoji{} & \addcheckemoji{}  \vspace{1pt}\\
        ReAct~\citep{Yao2022ReActSR}                    & \addcheckemoji{} & \addcrossemoji{} & \addcrossemoji{}              & \addcrossemoji{} & \addcheckemoji{} & \addcheckemoji{}  \vspace{1pt}\\
        Reflexion~\citep{shinn2023reflexion}            & \addcheckemoji{} & \addcrossemoji{} & \addcrossemoji{}              & \addcrossemoji{} & \addcheckemoji{} & \addcheckemoji{}  \vspace{1pt}\\
        $\dagger$ Self-Refine~\citep{madaan2023self_refine}       & \addcheckemoji{} & \addcheckemoji{} & \addcrossemoji{}              & \addcrossemoji{} & \addcheckemoji{} & \addcheckemoji{}  \vspace{1pt}\\
        Tree-of-thought~\citep{yao2023tree_of_thought}  & \addcheckemoji{} & \addcrossemoji{} & \addcrossemoji{}              & \addcrossemoji{} & \addcrossemoji{} & \addcheckemoji{}  \vspace{1pt}\\
        GPT-Bargaining~\citep{gpt-bargain}              & \addcrossemoji{} & \addcheckemoji{} & \addcheckemoji{} (fixed to 3) & \addcrossemoji{} & \addcheckemoji{} & \addcrossemoji{}  \vspace{1pt}\\
        Camel~\citep{li2023camel}                       & \addcheckemoji{} & \addcheckemoji{} & \addcheckemoji{} (fixed to 2) & \addcrossemoji{} & \addcheckemoji{} & \addcrossemoji{}  \vspace{1pt}\\
        ExpertPrompting~\citep{xu2023expertprompting}   & \addcheckemoji{} & \addcrossemoji{} & \addcrossemoji{}              & \addcheckemoji{} & \addcrossemoji{} & \addcheckemoji{}  \vspace{1pt}\\
        \textbf{\ours{} (ours)}                         & \addcheckemoji{} & \addcheckemoji{} & \addcheckemoji{} (varied)     & \addcheckemoji{} & \addcheckemoji{} & \addcheckemoji{}  \vspace{1pt}\\
    \bottomrule
\end{tabular}
}
\caption{High-level comparison with various prompting-based methods. Methods directly comparable to ours are denoted by $\dagger$. Results for the comparison can be found in Section~\ref{sec:exp}. In Section~\ref{sec:analysis}, we further design and compare with two variants of \ours{}: one adopting fixed personas, as in Camel~\citep{li2023camel}, and another with additional persona profiles, as proposed in ExpertPrompting~\citep{xu2023expertprompting}.
\vspace{-10pt}
}

\label{tab:comparison}
\end{table*}

To unleash the potential of cognitive synergy for general task-solving, we propose \textbf{\ours{}} \textbf{(\oursabbr)}, which \textit{prompts a single LLM to identify, simulate, and collaborate with multiple personas}. Figure~\ref{fig:teaser_figure} provides a high-level overview of \oursabbr{}. 
Here, a persona can represent either a domain expert, such as a movie enthusiast, or a target audience, such as a ten-year-old child. 
Through the dynamic identification of various personas, we empower a single LLM to acquire diverse domain knowledge accurately without additional retrieval systems. 
By facilitating multi-turn self-collaboration, we enable self-revision and self-feedback from various perspectives without requiring additional agents.


In real-world scenarios, such as those in creative industries, there is often a need to incorporate diverse information from different domains. Figure~\ref{fig:main_illustration} presents a concrete example of how \oursabbr{} operates on a challenging task that requires creative integration of information from various domains, such as the Legend of Zelda game, Harry Potter movies, and Jay Chou's albums. 
Standard prompting fails to generate satisfactory output due to missing essential information and factual errors. In contrast, \oursabbr{} produces informative and coherent answers by automatically identifying expert personas and engaging in a multi-turn self-collaboration. In this process, the AI Assistant persona iteratively writes drafts of the story, solicits feedback from other participants, and revises accordingly.

To explore the prevalence of cognitive synergy in different LLMs, we apply \oursabbr{} to LLMs with varying scales and capabilities, including GPT-4, GPT-3.5-turbo, and Llama-13b-chat. Comparative results show that cognitive synergy only emerges in GPT-4 and not in less capable models. This draws an interesting analogy to human development, as children typically start engaging in role-playing at the age of 2 to 3~\cite{piaget1954construction}, but not earlier.    
In summary, the key contributions of this paper are as follows:
\begin{itemize}

\vspace{-3pt}
\item We investigate whether LLMs can leveraging cognitive synergy for general task-solving. We introduce \textbf{\ours{} (\oursabbr)}, which simulates multi-agent, multi-persona collaboration in a pure zero-shot manner.

\vspace{-3pt}
\item We evaluate \oursabbr{} across \textbf{three challenging tasks}: Trivia Creative Writing, Codenames Collaborative and Logic Grid Puzzle, spanning both knowledge- and reasoning-intensive domains. To our knowledge, \oursabbr{} is the first zero-shot prompting method that can enhance both knowledge and reasoning abilities on GPT-4.

\vspace{-3pt}
\item We present an intriguing finding regarding the emergent nature of cognitive synergy ability in LLMs, which \textbf{only emerges in GPT-4} and not in less powerful models.

\vspace{-3pt}
\item We conduct in-depth analyses of the impact of the identified personas and \oursabbr{} prompt design, providing insights into why \textbf{dynamic, fine-grained personas} are necessary, as opposed to fixed, coarse-grained personas.

\end{itemize}

\section{\ours{}}
\label{subsec:spp_procedure}


To unleash the power of synergizing different personas to tackle complex problems, we propose \ours{} (\oursabbr{}) which instructs a LLM to perform the following the procedure for general task-solving: \textbf{(1) Persona Identification}: Identify multiple participants with special personas (including a leader persona: AI Assistant) that are essential for solving the particular task. \textbf{(2) Brainstorming}: The participants share knowledge and provide suggestions on how to approach the task based on their own expertise. \textbf{(3) Multi-Persona Iterative Collaboration}: The leader persona, AI Assistant, proposes initial solutions, consults the other participants for feedback, and revise the answer iteratively. Figure~\ref{fig:main_illustration} shows a walking example of \oursabbr{} during inference. Next, we formally describe the \oursabbr{} procedure in detail.

Given an input sequence $x$ and a model $\mathcal{M}$, let a prompt (including demonstration examples) prepended to the input to be $p$ and the final output to be $y$. Denote an intermediate generation before generating the final $y$ as $z$. Under this formulation, Standard Prompting and Chain-of-Thought (CoT) Prompting can be described as:
\begin{align}
    &\scalebox{0.8}{\textit{Standard Prompting:} \quad{} $y = \mathcal{M}(x)$}\\
    &\scalebox{0.8}{\textit{CoT Prompting:} \quad{} $y = \mathcal{M}(p_{cot}\lVert x \lVert \{z_1,z_2,...,z_n\})$}
\end{align}
where $p_{cot}$ is the CoT prompt, e.g., \texttt{"Solve the task step-by-step"} and $\{z_1,z_2...,z_n\}$ are the intermediate steps. In contrast, our proposed \ours{} can be described as follows:
\begin{align}
    &\scalebox{0.8}{\textit{\ours{}:} \quad{} $y =$ } \nonumber \\
    &\scalebox{0.8}{$\mathcal{M}(p_{spp}\lVert x \lVert z_p \lVert \{z^1_b, z^2_b,...,z^m_b\} \lVert \{z^0_s, z^1_f,...,z^m_f\}_{j=1..n})$} \label{eq:spp}
\end{align}
where the \oursabbr{} prompt ($p_{spp}$) includes a high-level instruction and two carefully crafted demonstration examples\footnote{The tasks we use in the demonstration examples do not overlap with the evaluation tasks.} that showcase the expected task-solving procedure of \oursabbr{}. We describe the design details of the prompt in \textbf{\S\ref{subsec:spp_prompt_design}}.
The corresponding intermediate generations ($z$) of \oursabbr{} are detailed below. 

\vspace{-3pt}
\paragraph{Persona Identification ($z_p$).} Given an input task, \oursabbr{} first generates a list of participants with different personas. 
For example in Figure~\ref{fig:main_illustration}, the model identified a \textit{Jay Chou Fan} persona to help answer "the last song in the second album by Jay Chou".   
%
We let the language model identify the personas dynamically instead of manually defining them. Given only two demonstration examples (detailed in \S\ref{app:prompts}), we observe that a state-of-the-art large language model, e.g., GPT-4~\citep{openai2023gpt4}, can identify accurate and meaningful personas for diverse tasks. We denote this part of intermediate generation as $z_p$ in Equation~\ref{eq:spp}.   

\vspace{-3pt}
\paragraph{Brainstorming ($z^i_b$).}  
Among the identified participants, "AI Assistant (you)" is treated as a leader persona that initiates the collaboration and generates initial solutions. Before generating the initial answer, the personas brainstorm on how to approach the task from their own perspectives. For example, the \textit{Jay Chou Fan} points out that the last song in Jay Chou's second album is "An Jing" ("Silence").
We find that the brainstorming phase effectively improves the quality of the initial solution.   
In Equation~\ref{eq:spp}, the superscript $i=0$ is used to denote the "AI Assistant" persona, while $i\geq1$ represents other dynamically identified personas. The intermediate generations of the brainstorming step are denoted as $\{z^1_b, z^2_b,...,z^m_b\}$.  

\vspace{-3pt}
\paragraph{Multi-Persona Iterative Collaboration ($z^0_s$, $z^i_f$).}  
Based on the brainstorming remarks, the AI Assistant persona generates an initial solution $z^0_s$, then it consults each of the other participants for feedback $\{z^i_f\}$. The participants are encouraged to critique the current generation and give revision suggestions. For example, the Jay Chou Fan persona checks whether the song "An Jing" ("Silence") is correctly included in the story.
This process can be repeated for multiple times until every participant is satisfied with the current solution. 
In Equation~\ref{eq:spp}, we denote the intermediate generations of the multi-turn dialogue as $\{z^0_s, z^1_f,...,z^m_f\}_{j=1...n}$ where $n$ is the number of iterations before reaching the final answer. The final answer can be directly read out following user-specified output format.

In summary, \oursabbr{} instructs an LLM to solve general tasks via multi-persona self-collaboration in a pure zero-shot manner. In contrast, as detailed in Table~\ref{tab:comparison}, previous prompting-based methods are either task-specific or require additional mechanism, e.g., searching~\citep{yao2023tree_of_thought}, external tools~\cite{Yao2022ReActSR}, memory component~\citep{shinn2023reflexion}, and fine-tuning~\citep{xu2023expertprompting}.

\begin{table*}[th]
    \small
    \centering
\resizebox{\textwidth}{!}{
    \begin{tabular}{l | cc | cc || cc || cc}  
      \toprule  
      \multirow{2}{*}{\textbf{Methods}}
      & \multicolumn{2}{c|}{\textbf{Trivia.C.W (N=5)}}
      & \multicolumn{2}{c||}{\textbf{Trivia.C.W (N=10)}}
      & \multicolumn{2}{c||}{\textbf{Codenames.C}}
      & \multicolumn{2}{c}{\textbf{Logic.G.Puzzle}}
      \\
      & Score (\%) & $\Delta$
      & Score (\%) & $\Delta$
      & Score (\%) & $\Delta$
      & Score (\%) & $\Delta$
      \\
      \midrule  
          Standard & 74.6 & 0.0\% & 77.0 & 0.0\% 
          & 75.4   & 0.0\%
          & 57.7  & 0.0\% 
          \\  
          CoT & 67.1 & \textcolor{textred}{$\downarrow$10.0\%} & 68.5 & \textcolor{textred}{$\downarrow$11.1\%} 
          & 72.7   & \textcolor{textred}{$\downarrow$3.6\%}
          & 65.8  & \textcolor{textgreen}{$\uparrow$14.1\%}
          \\  
          \midrule
          Self-Refine [iter=0] & 73.8 & & 76.3 &  
          & 75.2   & 
          &   58.8  & 
          \\
          Self-Refine [iter=1] & 73.9 & \textcolor{textred}{$\downarrow$1.0\%} & 76.9 & \textcolor{textred}{$\downarrow$0.1\%} 
          & 64.6   & \textcolor{textred}{$\downarrow$14.6\%}
          &   60.0  & \textcolor{textgreen}{$\uparrow$4.0\%} 
          \\  
          \midrule
          \textbf{SPP (ours) } & \textbf{79.9}   & \textcolor{textgreen}{\textbf{$\uparrow$7.1\%}} & \textbf{84.7}   & \textcolor{textgreen}{\textbf{$\uparrow$10.0\%}} 
          & \textbf{79.0}   & \textcolor{textgreen}{\textbf{$\uparrow$4.8\%}}
          &   \textbf{68.3}  & \textcolor{textgreen}{\textbf{$\uparrow$18.5\%}} 
          \\
      \bottomrule  
    \end{tabular}
}
  \caption{GPT-4 results on \tasktrivia{} (Trivia.C.W), \taskcodenames{} (Codenames.C) and \tasklogic{} (Logic.G.Puzzle). $\Delta$ indicates the relative gain/loss compared with Standard Prompting (first row). We report the average scores across two individual runs with/without a system message (detailed in Appendix~\ref{app:inference_config}).
  \vspace{-10pt}} 
  \label{tab:gpt4-results}  
\end{table*}

\section{Experiments}
\label{sec:exp}


To explore the effectiveness of \ours{} (\oursabbr{}), we adopt an evaluation methodology similar to that of previous work~\cite{yao2023tree_of_thought}. We carefully design new tasks and select tasks from existing benchmarks~\cite{bigbench} that are challenging even for the most capable LLMs~\citep{openai2023gpt4}.
The evaluation aims to cover diverse types of tasks encompassing both \textit{knowledge-intensive} and \textit{reasoning-intensive} domains. 

\vspace{-3pt}
\paragraph{Tasks.} We invent the \textbf{\tasktrivia{}} task (\S\ref{subsec:trivia_task}), which requires the model to internally acquire and integrate diverse information from various fields. 
We observe that even GPT-4~\citep{openai2023gpt4} frequently exhibit hallucination and factuality errors in the \tasktrivia{} task. 
We also propose the \textbf{\taskcodenames{}} task (\S\ref{subsec:codenames_task}), an extension of the Codenames task from the BigBench~\citep{bigbench} that features a two-role collaboration setup. 
\taskcodenames{} demands creative reasoning across a broad range of related knowledge and challenges the model's theory of mind skills. 
Lastly, we include a challenging pure-reasoning task, \textbf{Logic Grid Puzzle} (\S\ref{subsec:logic_task}), from the BigBench~\citep{bigbench} which necessitates complex multi-step reasoning.

\vspace{-3pt}
\paragraph{Baselines.} We compare our approach with \textbf{Standard Prompting}, \textbf{Chain-of-Thought (CoT)} prompting methods (outlined in \S\ref{subsec:spp_procedure}) and \textbf{Self-Refine}~\citep{madaan2023self_refine}. 
For CoT, a similar prompt design to \cite{yao2023tree_of_thought} is employed, where the model is prompted to generate a plan or a series of steps before producing the final output.
For Self-Refine, we follow \cite{madaan2023self_refine} to design \textit{feedback} and \textit{refine} prompts. We perform one self-refine iteration which requires three times more inferences than \oursabbr{}.
Full prompts for the methods can be found in Appendix~\ref{app:subsec:full_prompts}.

\vspace{-3pt}
\paragraph{Models.} The default model we use is GPT-4~\citep{openai2023gpt4}. 
Detailed inference configurations, API versions, and full results can be found in Appendices~\ref{app:inference_config} and \ref{app:full_results}. 
In \S\ref{subsec:model_comparision_emergent_ability}, we further investigate the prevalence of cognitive synergy in LLMs with different scales and capabilities, including GPT-3.5-turbo~\citep{gpt35} and Llama2-13b-chat~\citep{llama2}.



\begin{figure*}[t]
  \centering
  \includegraphics[width=0.75\textwidth]{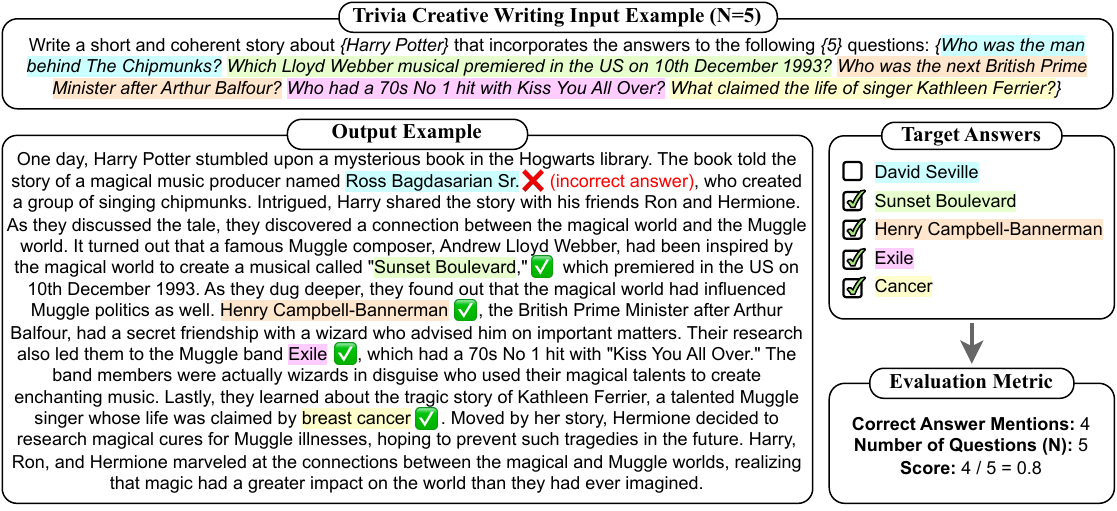}
  \caption{
  \tasktrivia{} task example.
  }
  \label{fig:trivia_task}
\end{figure*}
\begin{figure*}[t]
  \centering
  \includegraphics[width=0.75\textwidth]{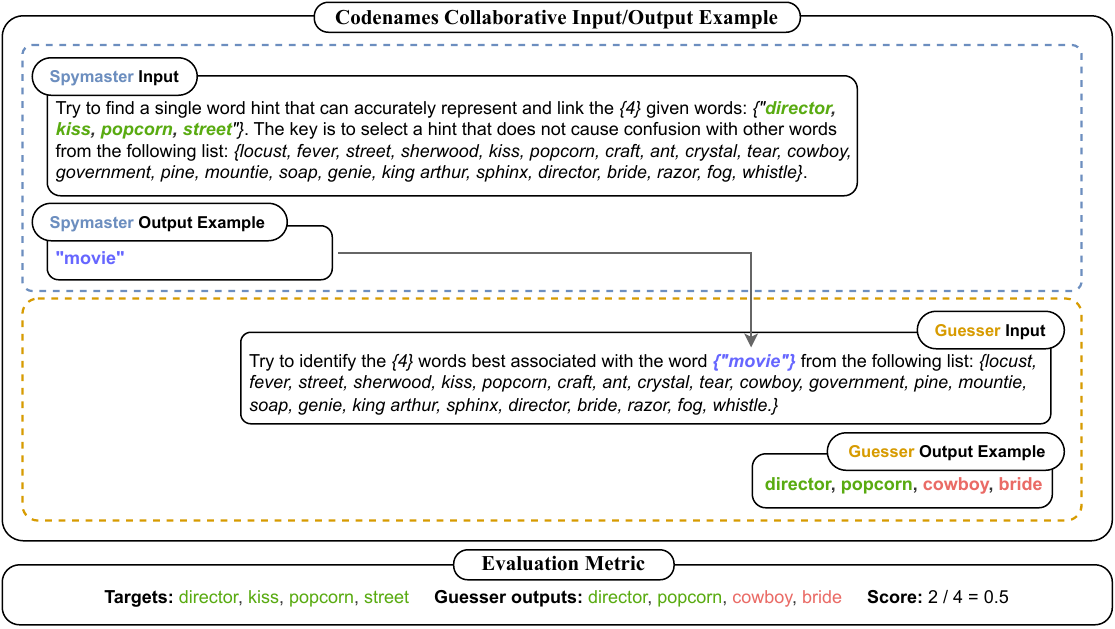}
  \caption{
  \taskcodenames{} task example.
  }
  \label{fig:codenames_task}
\end{figure*}
\begin{figure*}[t]
  \centering
  \includegraphics[width=0.75\textwidth]{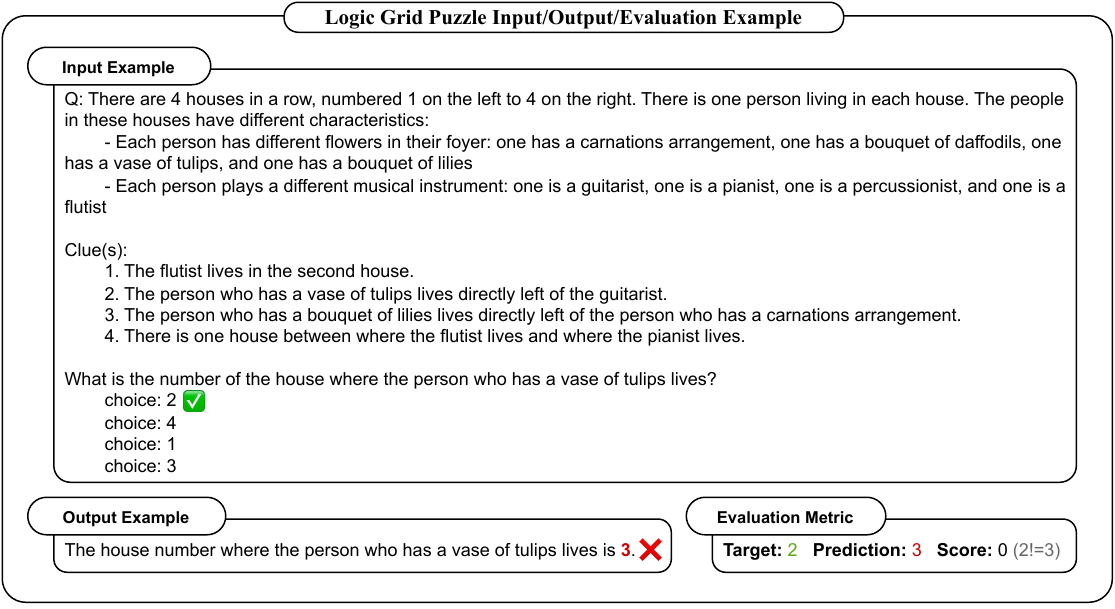}
  \caption{
  \tasklogic{} task example.
  }
  \label{fig:logic_task}
\end{figure*}

\subsection{\tasktrivia{}: A Knowledge-Intensive Task}
\label{subsec:trivia_task}

\paragraph{Task Description.}
%
As illustrated in Figure~\ref{fig:trivia_task}, \tasktrivia{} asks a model to write a coherent story while incorporating the answers to $N$ trivia questions. Our preliminary experiments (Figure~\ref{fig:TCW_analysis_N_shuffle}) show that a sufficiently large $N$ can effectively challenge GPT-4 to demonstrate factual knowledge across diverse domains. Thus, we mainly consider two evaluation settings, $N=5$ and $N=10$. 
We built a benchmark with 100 instances for each $N$, covering a total of 1000 trivia questions\footnote{To select difficult question instances that can pose challenges to GPT-4, we use a smaller open-source LLM, \textit{fastchat\_t5\_3b}~\citep{zheng2023judging}, to obtain preliminary performance on the validation set, and then choose the failure cases as our question selection.} extracted from the TriviaQA~\citep{joshi-etal-2017-triviaqa} dataset. 
More details can be found in Appendix~\ref{app:task_detail_trivia}.

\vspace{-3pt}
\paragraph{Evaluation Metrics.} 
Evaluating GPT-4 level generation results can be challenging. Our preliminary experiments indicate that, even for humans, it is very difficult to identify which generation is better in terms of overall "quality" of the story from different prompting methods. Thus, instead of focusing on evaluating the coherence of the generation, which can be highly subjective, we employ an automatic metric which focuses on detecting factual hallucinations. 
As shown in Figure~\ref{fig:trivia_task}, we perform string matching with the ground truth target answers for each question on the output generation. 
For each question, a match to any of the answer aliases provided by the TriviaQA dataset is considered a correct mention. 
The metric score is computed as: $\frac{\text{\# correct answer mentions}}{\text{\# trivia questions}}$.


\vspace{-3pt}
\paragraph{Results.}
Table~\ref{tab:gpt4-results} presents the results of the \tasktrivia{} task. The key observations are as follows: (1) Chain-of-Thought (CoT) does not outperform Standard prompting, indicating that CoT is ineffective in eliciting an LLM's knowledge abilities. Qualitative examples in Figure~\ref{fig:qaulitative_short} and \ref{fig:trivia-spp-vs-cot-qualitative} illustrate that although CoT generates reasonable plans for task resolution, the final generation still contains factual errors and hallucinations. (2) Self-Refine only brings marginal improvements over iterations. (3) \oursabbr{} outperforms all baselines significantly. The improvement is more pronounced in the $N=10$ setting compared to $N=5$ (10\% vs. 7\%), suggesting that \ours{} is particularly beneficial when the task requires incorporating knowledge from numerous domains.

\subsection{\taskcodenames{}: A Knowledge+Reasoning Task}
\label{subsec:codenames_task}

\paragraph{Task Description.} 
As illustrated in \ref{fig:codenames_task}, \taskcodenames{} is a collaborative task that challenges a model's knowledge, reasoning, and theory of mind abilities by assigning two player roles: the \textit{Spymaster} and the \textit{Guesser}. The Spymaster's role is to provide a hint word related to the target words, excluding some other distractor words, while the Guesser's role is to identify the target words based on the given hint and the full list of words. 
The same LLM (GPT-4~\citep{openai2023gpt4}) is used for both roles sequentially, and a dataset with 50 instances is constructed based on BigBench's~\citep{bigbench} Codenames task data.

\vspace{-3pt}
\paragraph{Evaluation Metrics.} The original Codenames task in the BigBench dataset has limitations due to its focus on the Guesser role and subjectivity in hint words. Our new task, \taskcodenames{}, resolves this by creating a self-contained evaluation setting that accurately measures the model's capability without human annotation. As illustrated in Figure~\ref{fig:codenames_task}, we compute the overlapping ratio between the predicted words from the Guesser and the target words as the metric. 



\vspace{-3pt}
\paragraph{Results.}
Table~\ref{tab:gpt4-results} shows the results on the \taskcodenames{} task. Similar to the \tasktrivia{} task, we find that CoT does not bring positive gains compared with the Standard prompting. 
Interestingly, iterative self-refinement brings negative impact on this task, due to a high tendency changing the initial response even if it is already good.
In contrast, \oursabbr{} brings significant improvements (\textasciitilde{}5\%), which indicates its effectiveness on collaborative tasks that require knowledge, reasoning, and theory of mind skills. 
Figure~\ref{fig:codenames-spp-vs-cot-qualitative} provides further qualitative examples illustrating that \oursabbr{} generates \textit{detailed} and \textit{interpretable} intermediate dialogues.

\subsection{\tasklogic{}: A Reasoning-Intensive Task}
\label{subsec:logic_task}

\paragraph{Task Description and Evaluation Metrics}
We utilize the \tasklogic{} task from the Bigbench~\citep{bigbench} dataset, which comprises 200 instances. Each instance describes a logic puzzle typically involving 2 to 5 houses, with each house inhabited by a person with specific characteristics, such as playing the piano. The objective is to answer questions about house numbers based on given clues, which requires multi-step reasoning and the selection of relevant information. An example input and output of the \tasklogic{} task are illustrated in Figure~\ref{fig:logic_task}. For evaluation metrics, we calculate the accuracy of the predicted house numbers by comparing them with the ground truth targets provided by the dataset.


\vspace{-3pt}
\paragraph{Results.}

Table~\ref{tab:gpt4-results} presents the results on \tasklogic{}. In contrast to the previous two tasks, we find that CoT brings significant improvements compared to Standard prompting, verifying the observation from previous work that CoT elicits better reasoning abilities. Furthermore, we discover that \oursabbr{} also achieves strong performance on this reasoning-intensive task. 

\begin{figure*}[!ht]
  \centering
  \includegraphics[width=0.9\textwidth]{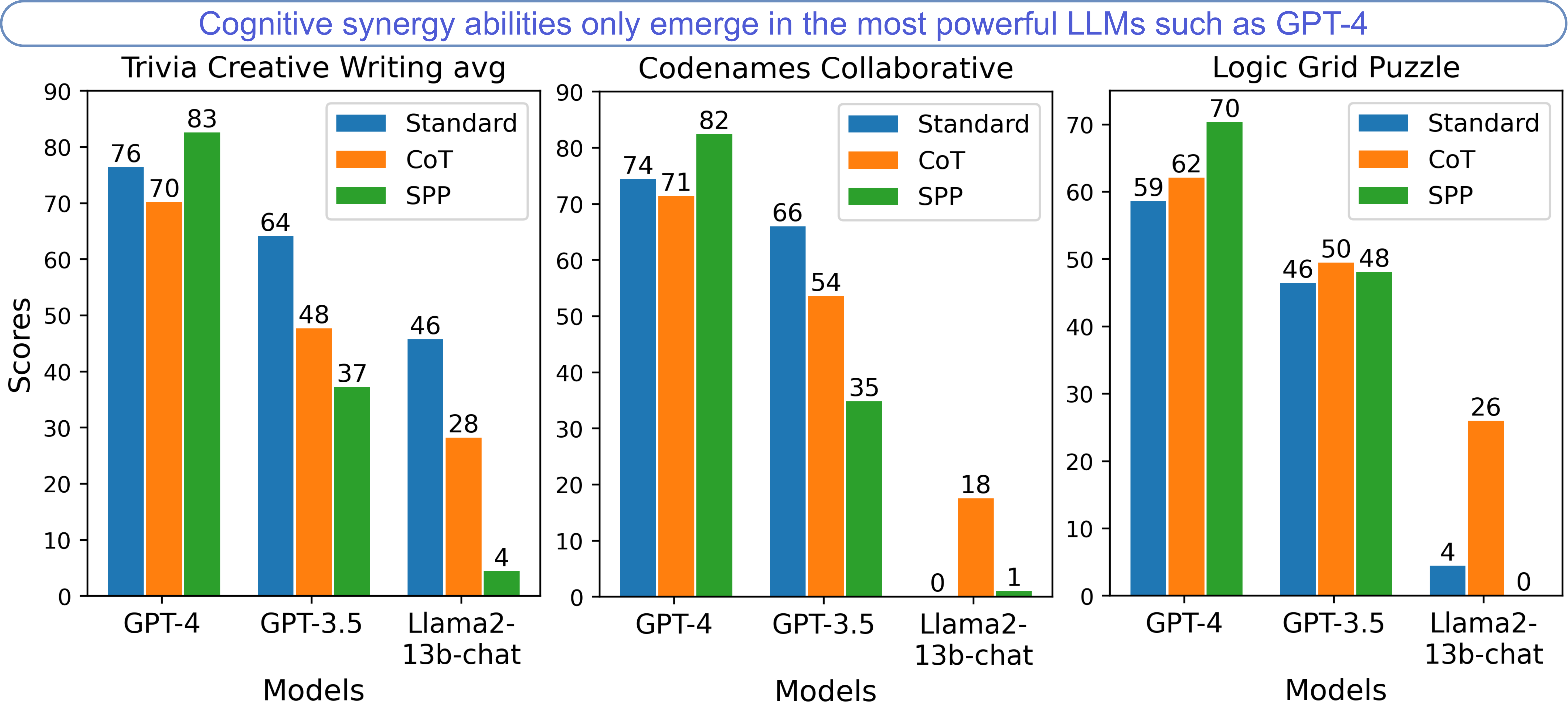}
  \vspace{-5pt}
  \caption{
    \oursabbr{} achieves superior performance only with the most powerful LLM (GPT-4), but not with GPT-3.5 and Llama2-13b. This indicates that cognitive synergy abilities only emerge in LLMs with GPT-4 level capabilities.
    \vspace{-10pt}
  }
  \label{fig:model_comparison}
\end{figure*}

\subsection{The Emergence of Cognitive Synergy}
\label{subsec:model_comparision_emergent_ability}

We further discover that \textbf{cognitive synergy can only be fully unleashed in LLMs with a certain level of instruction-following capabilities, akin to that of GPT-4.} This can be intriguingly compared to human development, where children usually begin to participate in role-playing around the ages of 2 to 3~\cite{piaget1954construction}, but not before that age.

As shown in Figure~\ref{fig:model_comparison}, the effectiveness of \oursabbr{} is not seen in smaller and less capable models like GPT-3.5 and Llama2. 
Additionally, on Llama2, we identify a unique problem which we refer to as \textit{early-termination}, where the model stops generating after identifying the participants, resulting in exceptionally low performance with \oursabbr{}. 
The model behaves as if it were waiting for input from a user instead of following the demonstration examples to generate responses on its own. Detailed discussions and examples on the early-termination problem can be found in Appendix~\ref{app:early-termination}. 




\begin{figure*}[t]
    \centering
    \begin{subfigure}[t]{0.22\linewidth}
        \begin{adjustbox}{valign=b}
            \includegraphics[width=\linewidth]{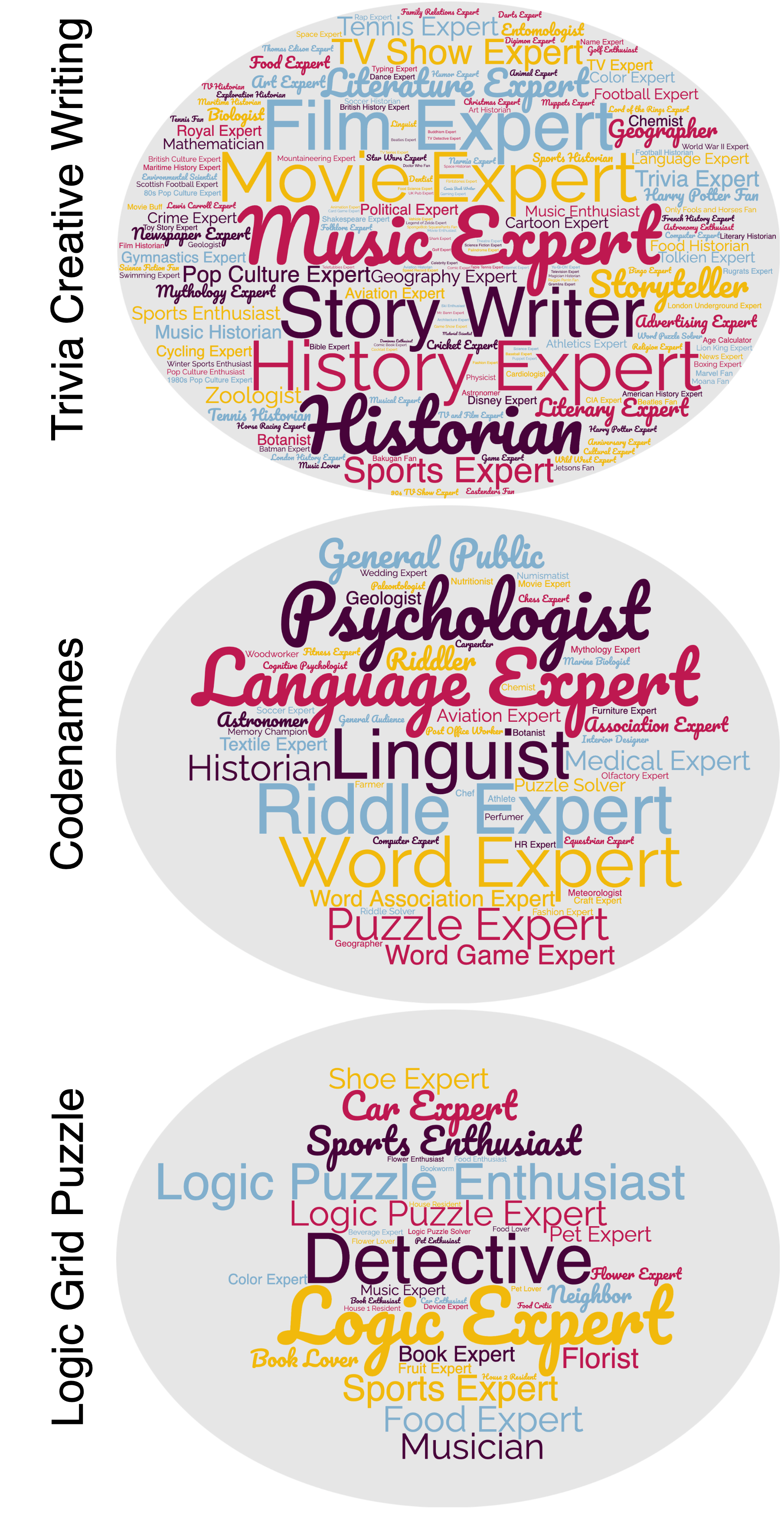}
        \end{adjustbox}
        \caption{Visualization of the \oursabbr{}-identified personas. The personas show a high correlation with the nature of the tasks.}
        \label{fig:analysis_merged_a}
    \end{subfigure}
    \hfill
    \begin{subfigure}[t]{0.69\linewidth}
        \begin{adjustbox}{valign=b}
            \includegraphics[width=\linewidth]{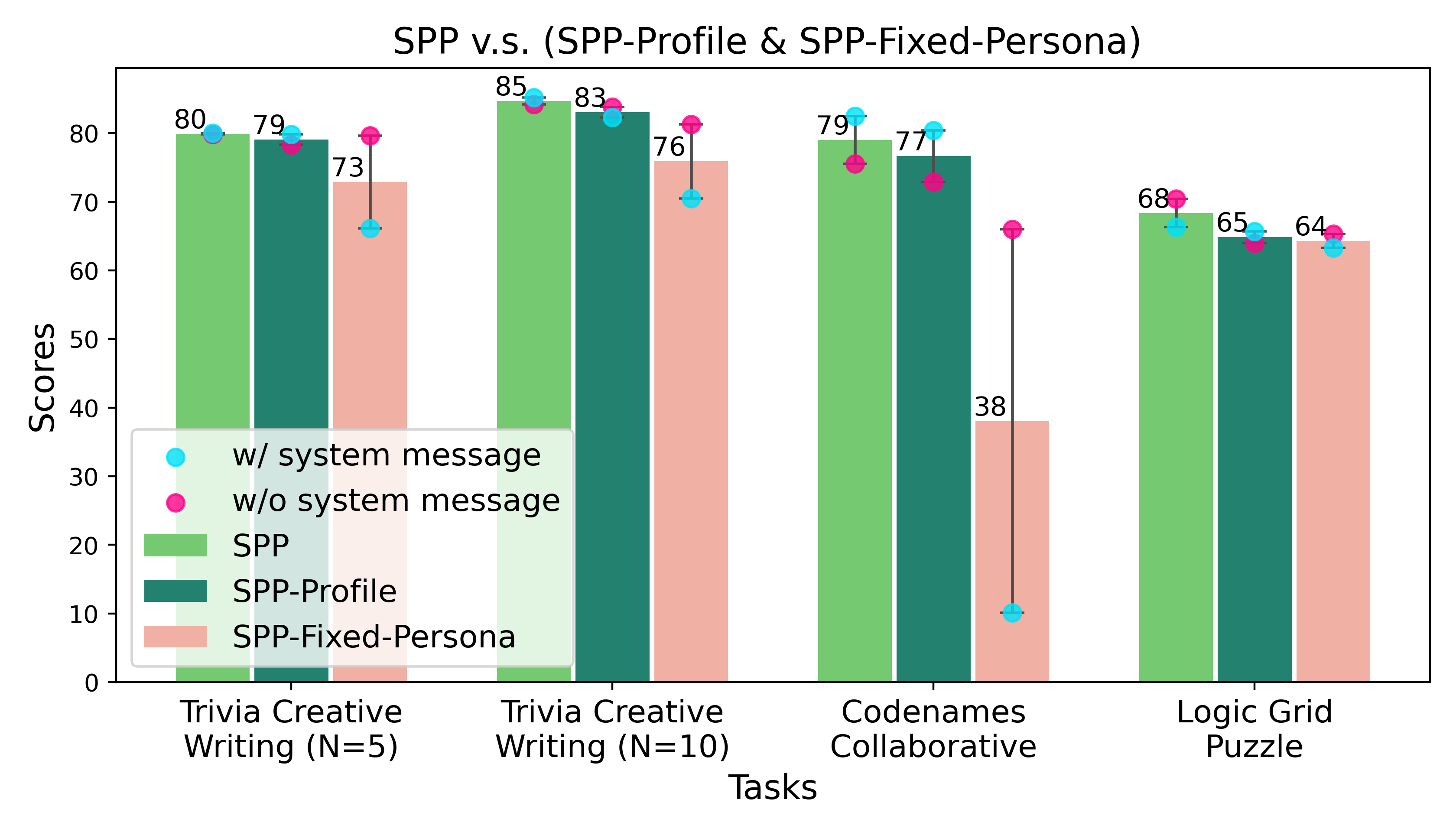}
        \end{adjustbox}
        \caption{ Comparison between \oursabbr{}, \oursfixabbr{} (with two fixed personas) and \oursproabbr{} (additionally generating persona profiles). \oursabbr{} significantly outperforms \oursfixabbr{}, highlighting the importance of automatically identifying dynamic, fine-grained personas. \oursabbr{} slightly outperforms \oursproabbr{}, indicating that the persona names (without detailed description of the expertise) are probably already sufficient for eliciting cognitive synergy.  
        }
        \label{fig:analysis_merged_b}
    \end{subfigure}
    \vspace{-6pt}
    \caption{\textbf{(a)} Qualitative analysis on the identified personas; \textbf{(b)} Quantitative analysis on two \oursabbr{} variants.}
    \label{fig:analysis_merged}
\end{figure*}

\vspace{-3pt}
\section{Analysis}
\label{sec:analysis}

\vspace{-3pt}
\paragraph{\oursabbr{} effectively improves both knowledge and reasoning abilities in LLMs.} As demonstrated by the results in \S\ref{sec:exp}, \ours{} (\oursabbr{}) not only brings significant improvements to knowledge-intensive tasks such as \tasktrivia{} and \taskcodenames{} without relying on external knowledge bases, but also achieves strong performance on reasoning-intensive tasks like \tasklogic{}. To our knowledge, \oursabbr{} is the first zero-shot prompting method that can enhance both knowledge and reasoning abilities on GPT-4.

\vspace{-3pt}
\paragraph{LLMs can effectively identify useful personas in a zero-shot manner.} 
We are interested in investigating whether the identified personas are highly relevant to the tasks.
We visualize the personas automatically identified by \oursabbr{} using a word cloud for each task in Figure~\ref{fig:analysis_merged_a}, where a larger font indicates a higher frequency. The key observations include: (1) The identified personas are closely correlated with the particular task. For example, in \tasklogic{}, even though "logic puzzle" is not mentioned in the input, the LLM frequently identifies the persona "Logic Puzzle Expert." 
(2) On knowledge-intensive tasks, such as \tasktrivia{}, \oursabbr{} identifies more diverse and specific personas, while on reasoning-intensive tasks, such as \tasklogic{}, the personas are more homogeneous.

We further investigate whether a detailed profile for each persona is needed for eliciting domain knowledge, as suggested by \cite{xu2023expertprompting}.
To this end, we design a variant of \oursabbr{}, \textbf{\oursproabbr{}}, which involves generating profiles for each persona during the Persona Identification phase. 
The results in Figure~\ref{fig:analysis_merged_b} show that \oursproabbr{} does not outperform \oursabbr{}. This suggests that a fine-grained persona name without a detailed description may already be sufficient for eliciting certain domain knowledge.



\begin{figure*}[!ht]
  \centering
  \includegraphics[width=0.85\textwidth]{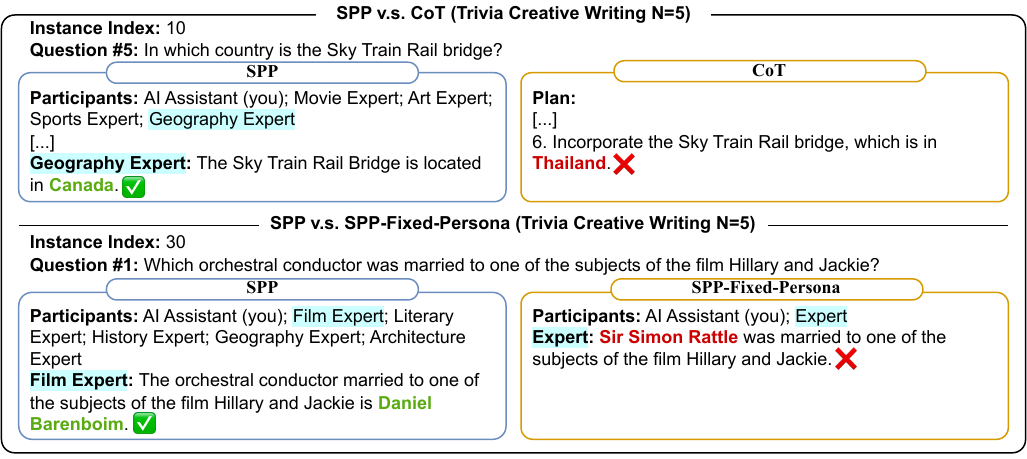}
  \caption{
  Qualitative examples on \tasktrivia{} comparing \oursabbr{}, CoT and \oursfixabbr{}. 
   While CoT provides reasonable intermediate steps, it still struggles with factual hallucination. \oursabbr{} v.s. \oursfixabbr{} reveals that dynamically identified fine-grained personas, such as the "Film Expert," tend to outperform the fixed general persona of an "Expert.
  More examples can be found in Figures~\ref{fig:trivia-spp-vs-cot-qualitative}, \ref{fig:codenames-spp-vs-cot-qualitative}, and \ref{fig:spp-fixed-qualitative}.
  \vspace{-10pt}
  }
  \label{fig:qaulitative_short}
\end{figure*}

\vspace{-4pt}
\paragraph{Dynamic personas v.s. fixed personas.} To further investigate the importance of dynamically identifying personas for each task instance instead of fixing a general persona, an ablated variant of \oursabbr{}, \textbf{\oursfixabbr{}}, is introduced. 
For \oursfixabbr{}, we modify the prompt (Figure~\ref{fig:full_prompts_spp_fixed}) to force the personas to be fixed as an "AI Assistant" and an "Expert".
%
Comparing \oursabbr{} and \oursfixabbr{} in Figure~\ref{fig:analysis_merged_b}, we have the following insights: (1) \oursabbr{} consistently outperforms \oursfixabbr{} across all tasks, suggesting that dynamic, fine-grained personas are more effective than fixed, general personas. Qualitative examples in Figure~\ref{fig:qaulitative_short} and \ref{fig:spp-fixed-qualitative} shows that the fine-grained personas such as "Film Expert" and "Sports Enthusiast" correctly provide the answers, while the fixed persona "Expert" fails.
(2) \oursfixabbr{} also suffers from the \textbf{early-termination} problem as defined in \S\ref{subsec:model_comparision_emergent_ability}, where the LLM stops collaboration before providing the final answer as if it were waiting for external inputs. 

\vspace{-3pt}
\paragraph{Impact of the demonstrations in \oursabbr{} prompt.}
To investigate the effectiveness of the hand-crafted demonstration examples in \oursabbr{}, we conduct an ablation study where we remove the second demo example and preserve the first one, which shows only a two-persona collaboration setting. As shown in Figure~\ref{fig:impact_demo_example}, we observe that (1) Adding the second example, which requires collaboration of more than two personas, effectively boosts the performance. (2) \oursabbr{} is fairly robust to the prompt change and show good performance with only the first demo example.

\vspace{-3pt}
\section{Related Work}
\vspace{-3pt}

\paragraph{LLMs as role-playing agents.}
Recent research~\citep{deshpande2023toxicity, xu2023expertprompting, gpt-bargain, autogpt, li2023camel} demonstrates that assigning personas or roles to LLMs influences their generation behavior. AI societies with distinct personas or occupations have been explored for collaboration~\citep{generative_agents, schick2022peer, li2023camel, toolmakers}. However, limitations in persona assignment and multi-agent collaboration include single or fixed persona assignments~\citep{xu2023expertprompting, gpt-bargain, schick2022peer, li2023camel} and the need for multiple LLM instances, increasing inference cost. In contrast, \oursabbr{} uses a single LLM to dynamically identify useful personas for general tasks. 
Our discovery on the emergent nature of cognitive synergy
also aligns with related work~\citep{olausson2023demystifying}, which investigates the emergent ability of self-debugging in code generation.

\vspace{-3pt}
\paragraph{Enhancing reasoning and factual knowledge in LLMs.}
LLMs face challenges in complex knowledge-intensive tasks due to hallucination~\citep{maynez-etal-2020-faithfulness} and reasoning-intensive tasks due to the lack of human-like slow thinking~\citep{sloman1996empirical, kahneman2011thinking}. Approaches like Chain-of-Thought (CoT) and Self-Refinement encourage LLMs to solve tasks step by step or iteratively revise their answers~\citep{wei2023chainofthought,think-step-by-step,zhang2022automatic, fu2022complexity, xue2023rcot, yao2023tree_of_thought, madaan2023self_refine, shinn2023reflexion, gou2023critic, self-debug, inner-monologue, Yao2022ReActSR}. However, these methods do not necessarily reduce factual hallucination. Retrieval augmented LLMs~\citep{retro, atlas, zemi, retrieval_reduce_hallucination} enhance knowledge acquisition but do not improve reasoning abilities. We propose \ours{} (\oursabbr{}) to elicit both knowledge and reasoning abilities in LLMs, improving factuality while maintaining strong performance on pure-reasoning tasks. 

\vspace{-3pt}
\section{Conclusion}

\vspace{-4pt}
\ours{} unleashes the cognitive synergy abilities within powerful LLMs, significantly reducing factual hallucination while enhancing reasoning. 
The performance is assessed using newly proposed tasks, e.g., \tasktrivia{} and \taskcodenames{}, demonstrating superior results compared to Standard, CoT and Self-Refine. 
The discovery of the emergent nature of cognitive synergy on different LLMs draws interesting analogy to human development.

\newpage

\section*{Limitations}
Although \ours{} exhibits promising improvements in acquiring factually correct knowledge compared to Standard prompting, it has some limitations. For instance, even when a fine-grained persona is assigned, the answer may still be incorrect. It remains unclear to what extent assigning a persona can help enhance domain knowledge in a specific area. Dedicated diagnostic experiments and theoretical efforts are needed to quantify the impact of having a persona or not. 

Furthermore, we currently adopt an identical \oursabbr{} prompt with the same two demonstration examples for any given task inputs, which may be suboptimal. 
Future work investigating how to find better demonstration examples conditioned on each input could further improve the effectiveness of \oursabbr{}. 

Last but not least, if given sufficient computational budget, a natural variant of \oursabbr{} could extend to a \textit{multi-agent cognitive synergist} setup where a leader persona identifies several expert agents and forms a cabinet to collaboratively solve a task. The multi-agent setup allows for leveraging richer computation power, larger local memory, and more flexible human-computer interaction, which could be essential for deploying to real-world applications.

\section*{Acknowledgements}
We would like to express our gratitude to the anonymous reviewers for their insightful comments and suggestions. We would also like to thank our colleagues and fellow interns at Microsoft Research Asia for their valuable internal discussions and feedback. 
Zhenhailong Wang and Heng Ji are partially supported by U.S. DARPA ECOLE Program No. \#HR00112390060 and U.S. DARPA ITM Program No. FA8650-23-C-7316. The views and conclusions contained herein are those of the authors and should not be interpreted as necessarily representing the official policies, either expressed or implied, of DARPA, or the U.S. Government.

\bibliography{anthology,custom}

\newpage

\appendix

\section{Prompts}
\label{app:prompts}

\subsection{\oursabbr{} Prompt Design}
\label{subsec:spp_prompt_design}
To prompt an LLM to behave as a cognitive synergist that follows the expected task-solving procedure as mentioned in \S\ref{subsec:spp_procedure}, we carefully designed the structure of the \oursabbr{} prompt as follows. The full prompts can be found in \S~\ref{app:subsec:full_prompts}.\footnote{We use the same prompt for any arbitrary tasks.}

\paragraph{System Principle.} The first part of the prompt contains a high-level instruction: \texttt{"When faced with a task, begin by identifying the participants who will contribute to solving the task. Then, initiate a multi-turn collaboration process until a final solution is reached. The participants will give critical comments and detailed suggestions whenever necessary."}

\paragraph{Demonstration Examples.} Then, we include two manually crafted demonstration examples to showcase the expected task-solving behavior. The first example describes a \textit{Game of 24} task, where we only include two personas: an AI Assistant and a Math Expert. This task aims to provide an example of a \textit{reasoning-intensive task}, where the AI Assistant needs to propose multiple proposals, and the other participants need to give \textit{fine-grained feedback} on where the current solution went wrong and how to improve it. The second example describes a poem-writing task with \textit{diverse requirements}, including lexical constraints, semantic constraints, and audience awareness. This task aims to provide an example of a \textit{knowledge-intensive task}, where diverse personas are required to collaboratively solve the task. This example also demonstrates a case where it is important to assign a dedicated persona to the audience, e.g., a ten-year-old child.

\paragraph{Task Prefix.} The last part of the prompt reminds the model to \texttt{"identify the participants and collaboratively solve the following task step by step."} followed by task-specific format instructions and inputs.

\begin{figure}[th]
  \centering
  \includegraphics[width=0.45\textwidth]{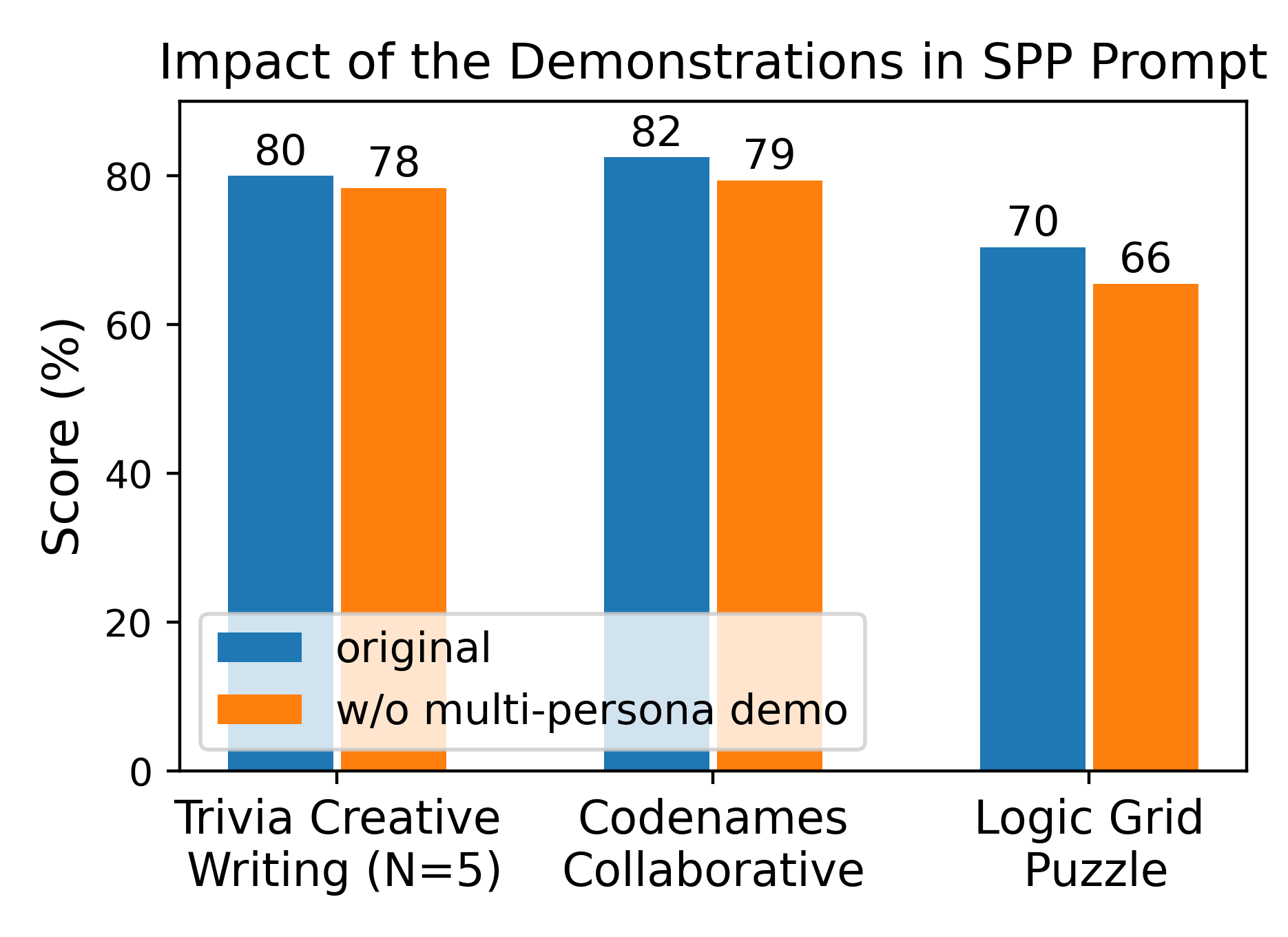}
  \caption{
      Analysis on the impact of the demonstration examples in \oursabbr{} prompt. We compare the effectiveness of the original \oursabbr{} prompt with a variant where we remove the second demonstration example, which shows a multi-persona scenario. We observe that (1) \oursabbr{} is fairly robust to the change in the prompt; (2) adding an additional multi-persona example apart from the single-persona one effectively boosts performance on all three tasks.
  }
  \label{fig:impact_demo_example}
\end{figure}


\subsection{Full Prompts}
\label{app:subsec:full_prompts}
\begin{figure*}[t]
  \centering
  \includegraphics[width=\textwidth]{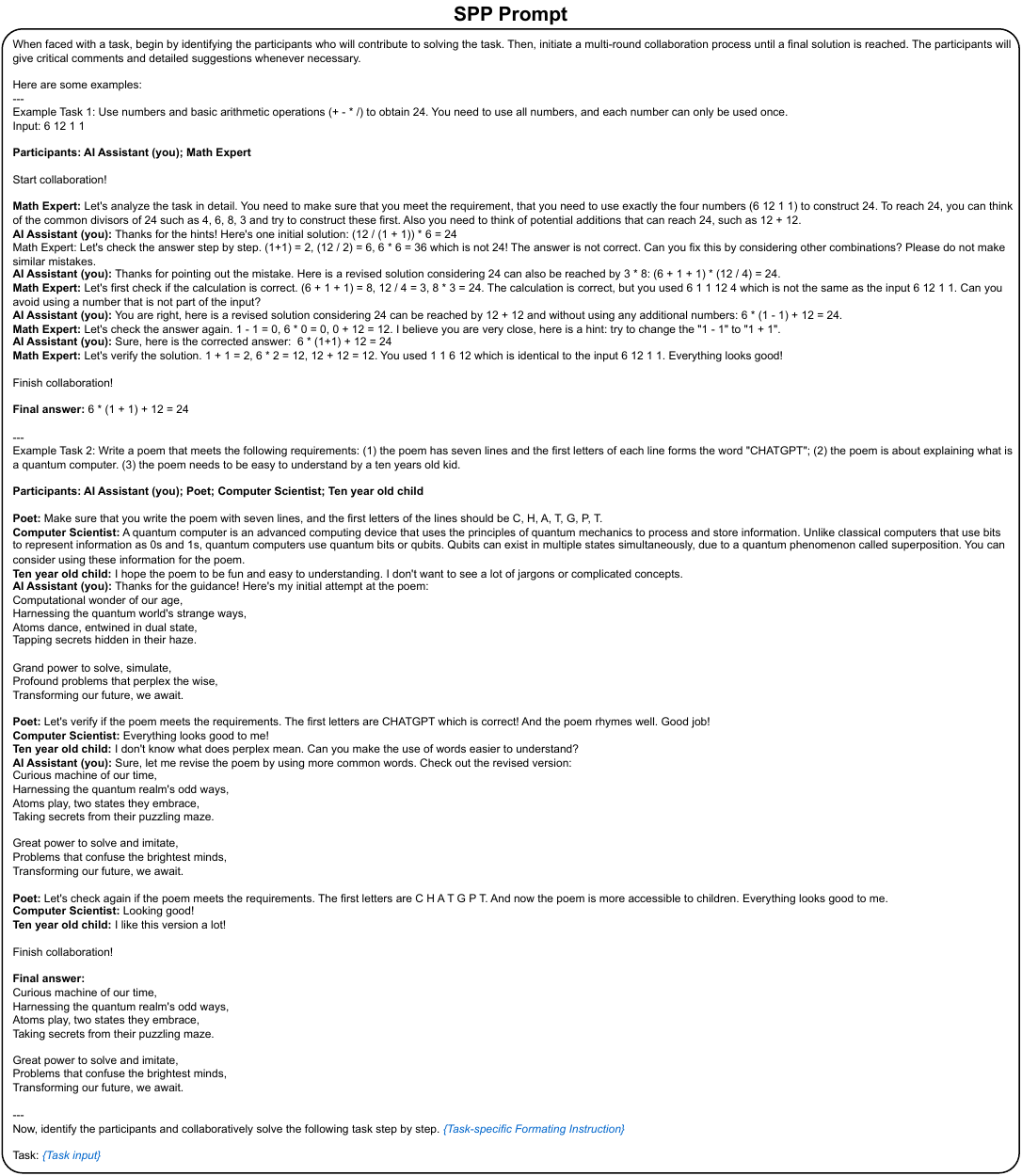}
  \caption{\oursabbr{} full prompt.
  }
  \label{fig:full_prompts_spp}
\end{figure*}

\begin{figure*}[t]
  \centering
  \includegraphics[width=\textwidth]{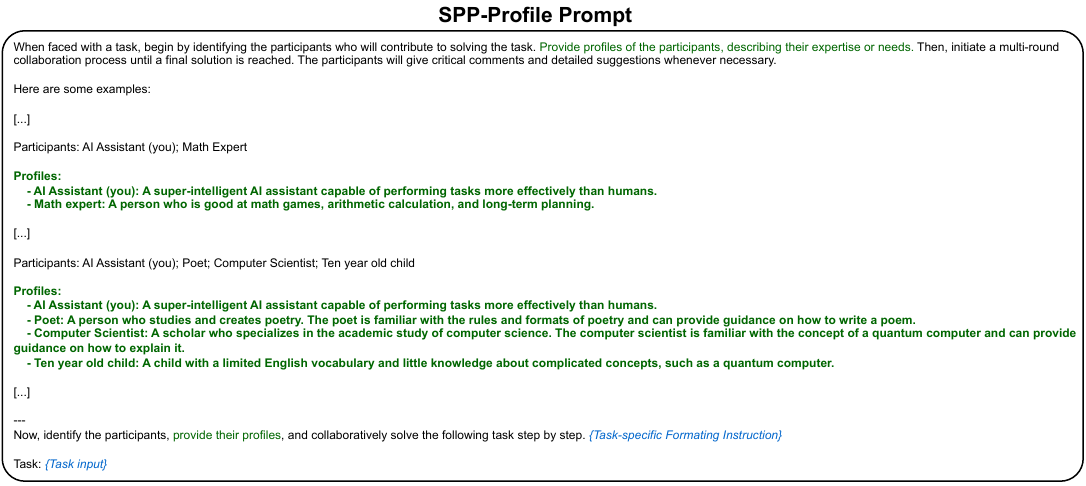}
  \caption{\oursproabbr{} full prompt. "[...]" indicates identical parts with \oursabbr{}. Green text indicates the key difference between \oursproabbr{} and \oursabbr{}.
  }
  \label{fig:full_prompts_spp_profile}
\end{figure*}

\begin{figure*}[t]
  \centering
  \includegraphics[width=\textwidth]{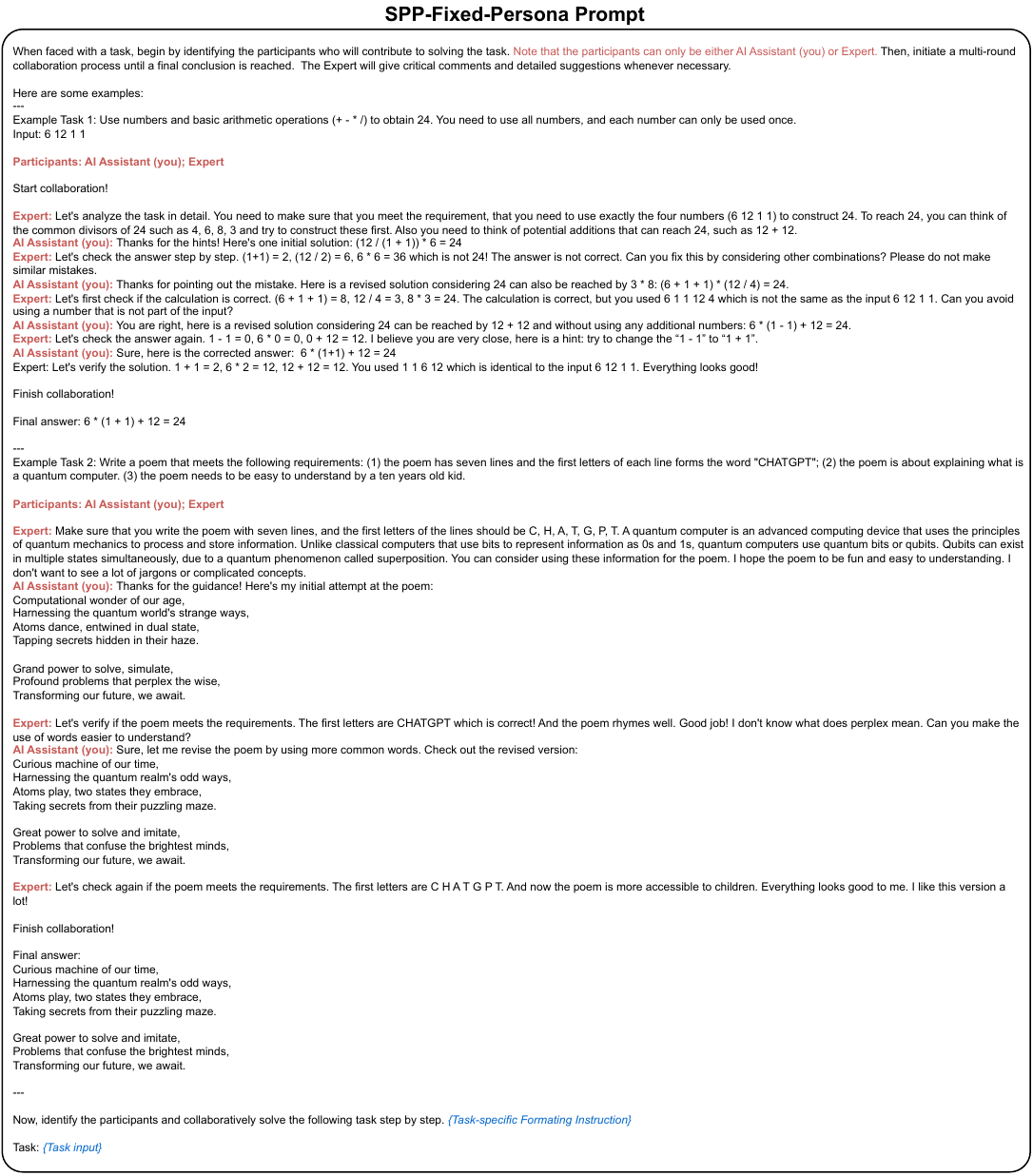}
  \caption{\oursfixabbr{} full prompt. Red text indicates the key difference between \oursfixabbr{} and \oursabbr{}.
  }
  \label{fig:full_prompts_spp_fixed}
\end{figure*}

\begin{figure*}[t]
  \centering
  \includegraphics[width=\textwidth]{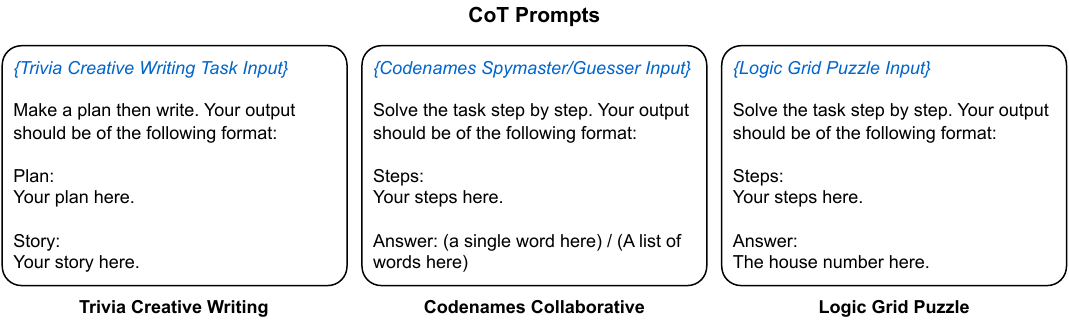}
  \caption{CoT prompts.
  }
  \label{fig:full_prompts_cot}
\end{figure*}

\begin{figure*}[t]
  \centering
  \includegraphics[width=\textwidth]{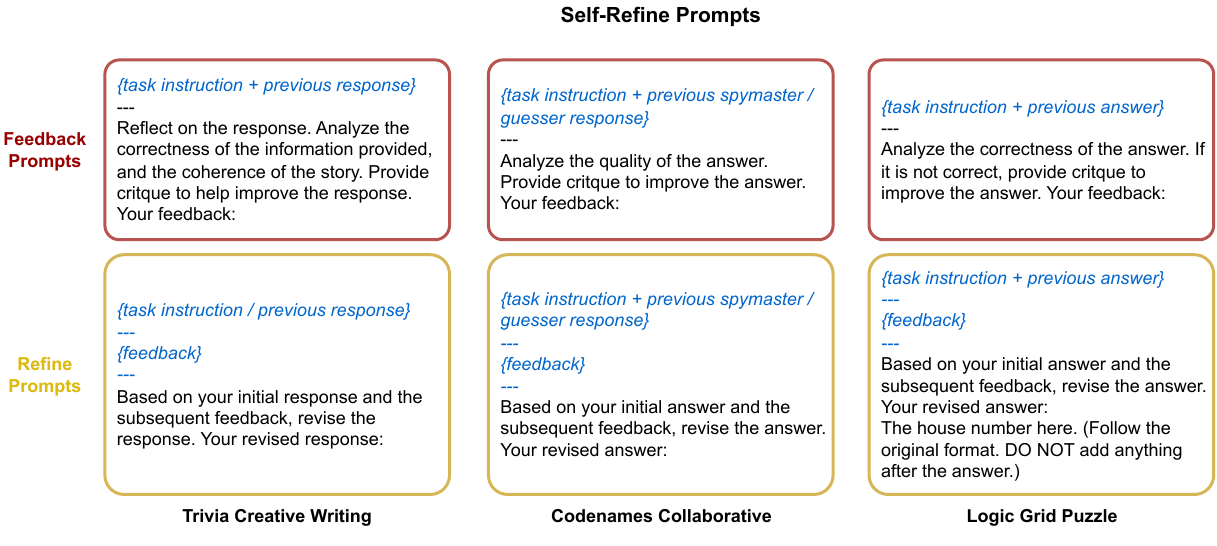}
  \caption{Self-refine prompts.
  }
  \label{fig:full_prompts_self_refine}
\end{figure*}

Figures~\ref{fig:full_prompts_spp}, \ref{fig:full_prompts_spp_profile} and \ref{fig:full_prompts_spp_fixed} show the full prompts for \oursabbr{}, \oursproabbr{} and \oursfixabbr{} respectively. Figure~\ref{fig:full_prompts_cot} shows the prompts for Chain-of-Thought (CoT) prompting. Figure~\ref{fig:full_prompts_self_refine} shows the prompts for Self-Refine prompting.

\section{Task Details}

\begin{figure*}[htb]
    \centering
    \begin{subfigure}[t]{0.45\linewidth}
        \begin{adjustbox}{valign=b}
            \includegraphics[width=\linewidth]{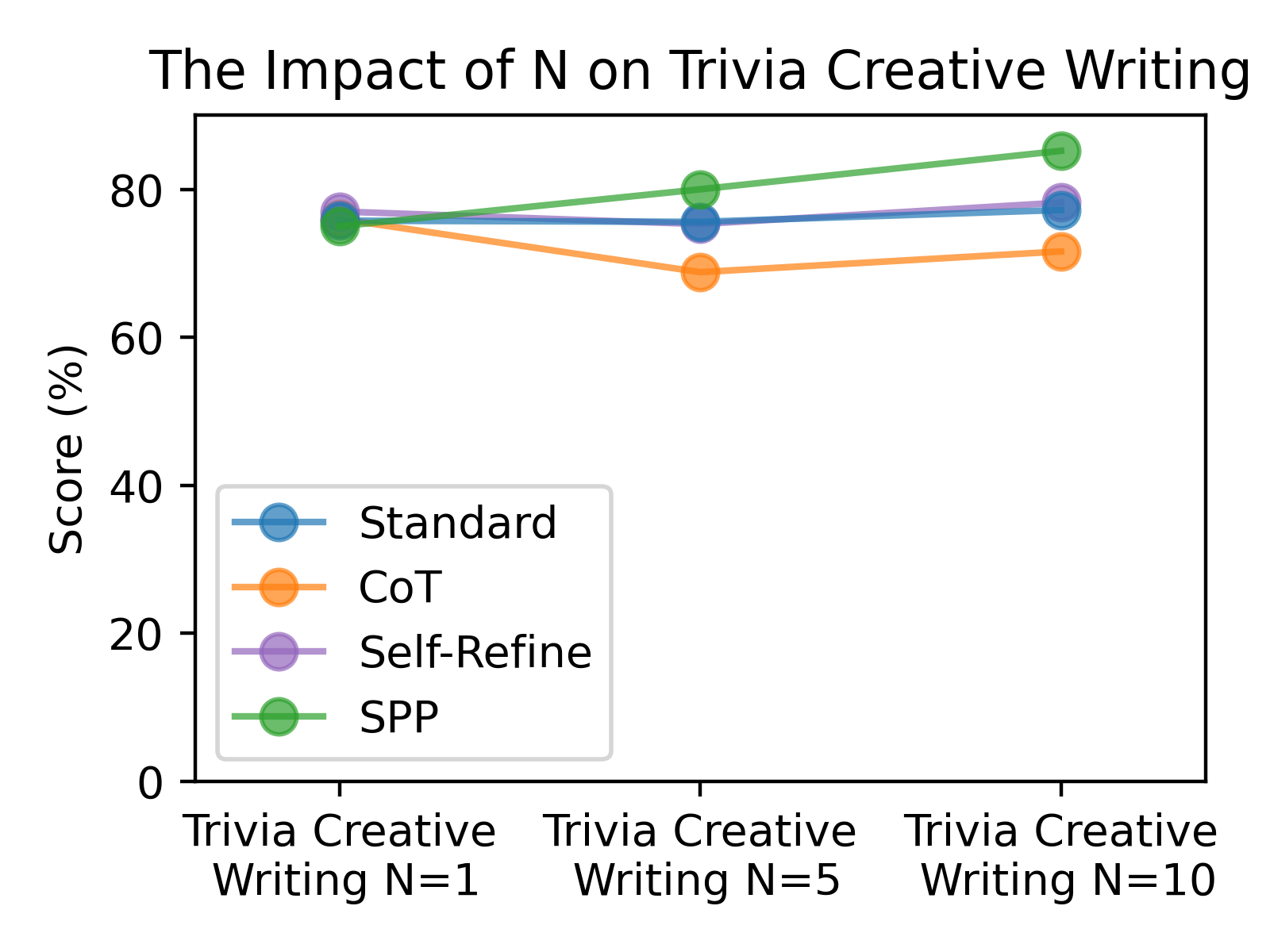}
        \end{adjustbox}
        \caption{Trivia Creative Writing with a large enough number of questions (N) effectively pose challenge to GPT-4 in terms of factual correctness. With N=1, different prompting methods result in similar performance, while with N>=5, \oursabbr{} shows visible superiority.}
    \end{subfigure}
    \hfill
    \begin{subfigure}[t]{0.45\linewidth}
        \begin{adjustbox}{valign=b}
            \includegraphics[width=\linewidth]{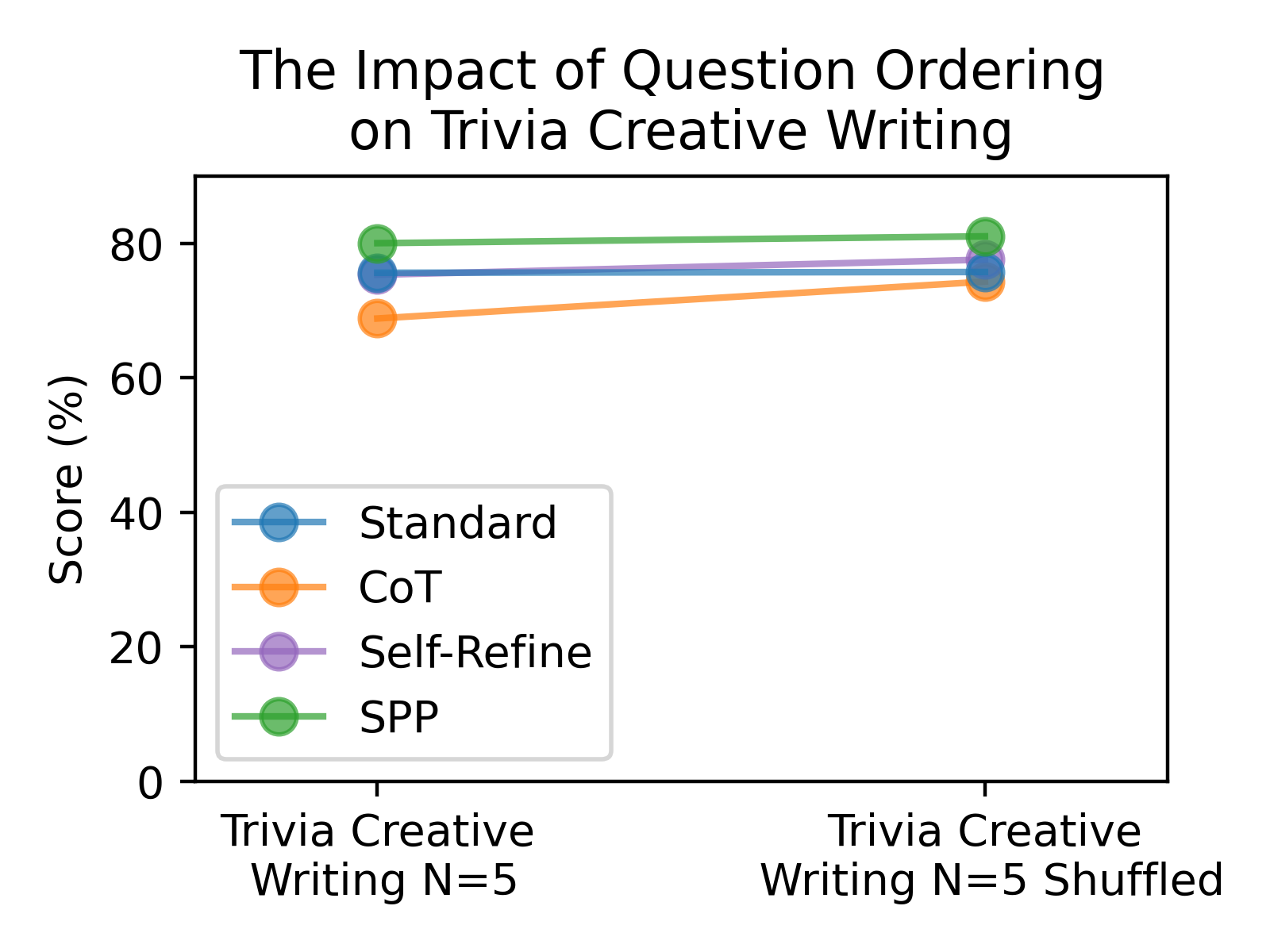}
        \end{adjustbox}
        \caption{The ordering of the questions in the Trivia Creative Writing task does not bring too much impact. The performance on shuffled questions is close to the original ordered questions.
        }
    \end{subfigure}
    \vspace{-5pt}
    \caption{Analysis on the impact of the number of questions (N) and the ordering of the questions for the Trivia Creative Writing task.}
    \label{fig:TCW_analysis_N_shuffle}
\end{figure*}

\subsection{\tasktrivia{}}
\label{app:task_detail_trivia}
Figure~\ref{fig:trivia_task} shows a detailed illustration of the \tasktrivia{} task. Additionally, we investigate how the number of the questions (N) and the ordering of the questions would affect the performance on the \tasktrivia{} task. As shown in Figure~\ref{fig:TCW_analysis_N_shuffle}, with a larger number of questions (N$\geq$5), \tasktrivia{} effectively challenges GPT-4's performance. While a single question (N=1) yields similar outcomes regardless of the prompting method, \oursabbr{} approach is notably superior for larger Ns. The ordering of the questions has minimal impact to the task performance. 

\begin{figure*}[t]
  \centering
  \includegraphics[width=\textwidth]{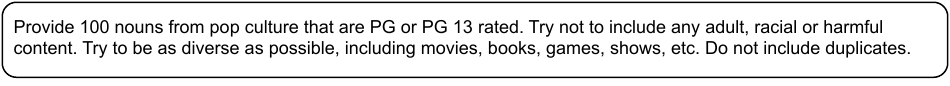}
  \caption{Prompt for generating the topic list for the \tasktrivia{} task.
  }
  \label{fig:topic_prompt}
\end{figure*}

The topic list is automatically generated by prompting GPT-4 to provide 100 nouns from pop culture\footnote{The full prompt for generating the topic list can be found in Figure~\ref{fig:topic_prompt}. We performed further human curation to avoid potential harmful content.}. 



\section{Inference Configurations}
\label{app:inference_config}
The main results in Table~\ref{tab:gpt4-results} are obtained from GPT-4. The GPT-4 API version we employ is Azure 2023-3-15-preview.\footnote{There are rare cases when a generation triggers the content filter of the API. We exclude those instances from our results.} The \textit{temperature} is set to $0.0$ (most conservative) and \textit{top\_p} to $1.0$ for all generations to maximize reproducibility. Since even though the temperature is set to $0.0$ the GPT-4 generation can still be non-deterministic, we conduct additional experiment to investigate its generation consistency under this configuration. As shown in Table~\ref{tab:gpt4_generation_consistency}, we perform three individual runs and compute the mean and standard deviation of the metric score on Trivia Creative Writing. We find that the variance is sufficiently small and \ours{} enjoys lower variance than Standard and CoT prompting.
\begin{table}[!ht]
    \centering  
    \small
    \begin{tabular}{l | c c c | c}  
      \toprule  
      \textbf{Methods} & \textbf{Run 1} & \textbf{Run 2 }&\textbf{ Run 3} & \textbf{Mean (std)}\\
      \midrule  
          Standard                                       & 75.6 & 74.4 & 73.1 & 74.4 \footnotesize $\pm 1.3$ \\
          CoT                                            & 68.8 & 69.6 & 70.8 & 69.7 \footnotesize $\pm 1.0$ \\
          \oursabbr{}                                    & 80.0 & 79.8 & 80.8 & 80.2 \footnotesize $\pm 0.5$ \\
      \bottomrule  
    \end{tabular}
    \caption{Investigation on the generation consistency of GPT-4 API. The experiment is performed on the Trivia Creative Task (N=5). We set the inference temperature to 0.0 and top\_p to 1.0 as all experiments conducted in the paper. The results show that the GPT-4 generation is fairly consistent with a small variance ($\sim 1\%$). We also observe that \oursabbr{} shows lower variance compared with Standard and CoT prompting across different runs.}
  \label{tab:gpt4_generation_consistency}  
\end{table}  

To evaluate the potential impact of initial persona assignment through a system message, we consider two inference settings: \textit{with} or \textit{without} the default system message, \texttt{"You are an AI assistant that helps people find information"}. Divergent patterns are observed across various tasks and methods regarding the use of the system message. We report the average metric scores across both inference settings in Table~\ref{tab:gpt4-results}. Full GPT-4 results for each setting can be found in Appendix~\ref{app:full_results}.

For GPT-3.5 results in Figure~\ref{fig:model_comparison}, we employ the same prompt, hyper-parameters and the best system message setting in terms of \oursabbr{}'s GPT-4 performance. For Llama2, we leverage the Huggingface text-generation pipeline\footnote{https://huggingface.co/blog/llama2} with greedy decoding.

\section{Additional Qualitative Analysis}
\begin{figure*}[t]
  \centering
  \includegraphics[width=\textwidth]{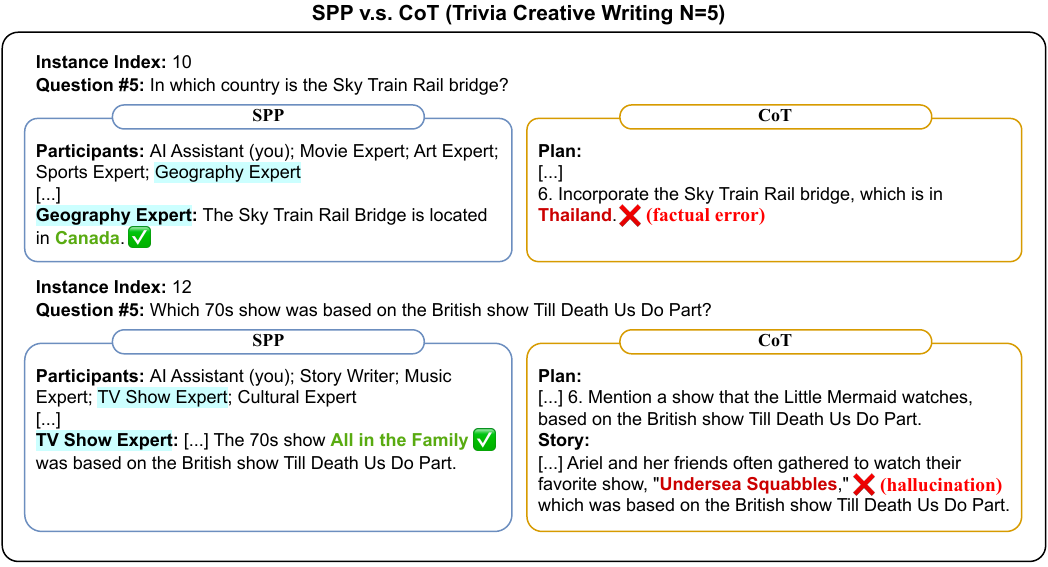}
  \caption{
  SPP vs CoT qualitative examples on \tasktrivia{} (N=5). We find that although CoT generates reasonable plans or steps, it tends to suffer from factual errors and hallucination.}
  \label{fig:trivia-spp-vs-cot-qualitative}
\end{figure*}
\begin{figure*}[t]
  \centering
  \includegraphics[width=\textwidth]{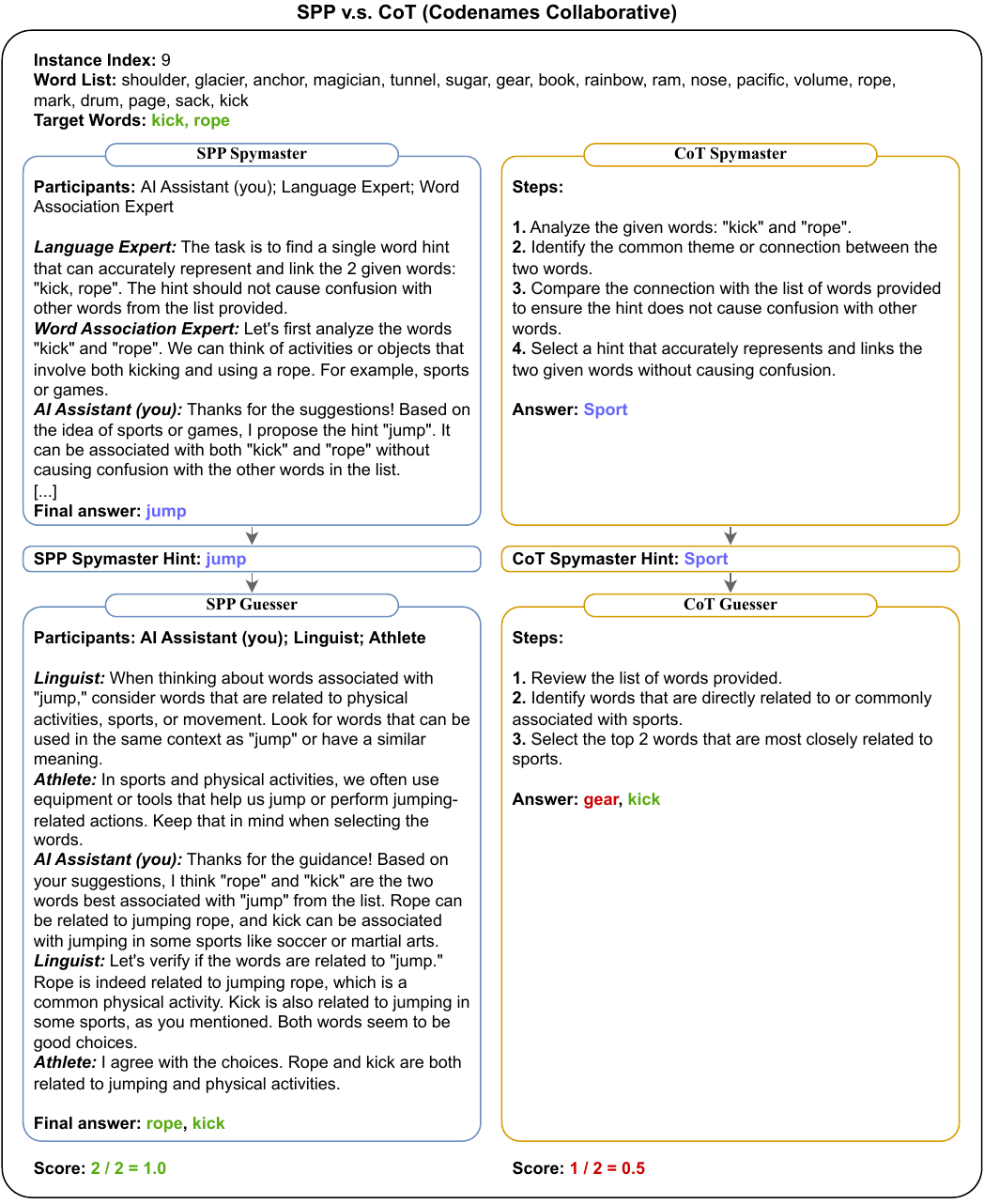}
  \caption{
  SPP vs CoT qualitative examples on \taskcodenames{}. We find that \oursabbr{} provides much more detailed and interpretable intermediate discussions from various perspectives, which leads to stronger knowledge selection, integration, and theory-of-mind capabilities.
  }
  \label{fig:codenames-spp-vs-cot-qualitative}
\end{figure*}
\begin{figure*}[t]
  \centering
  \includegraphics[width=\textwidth]{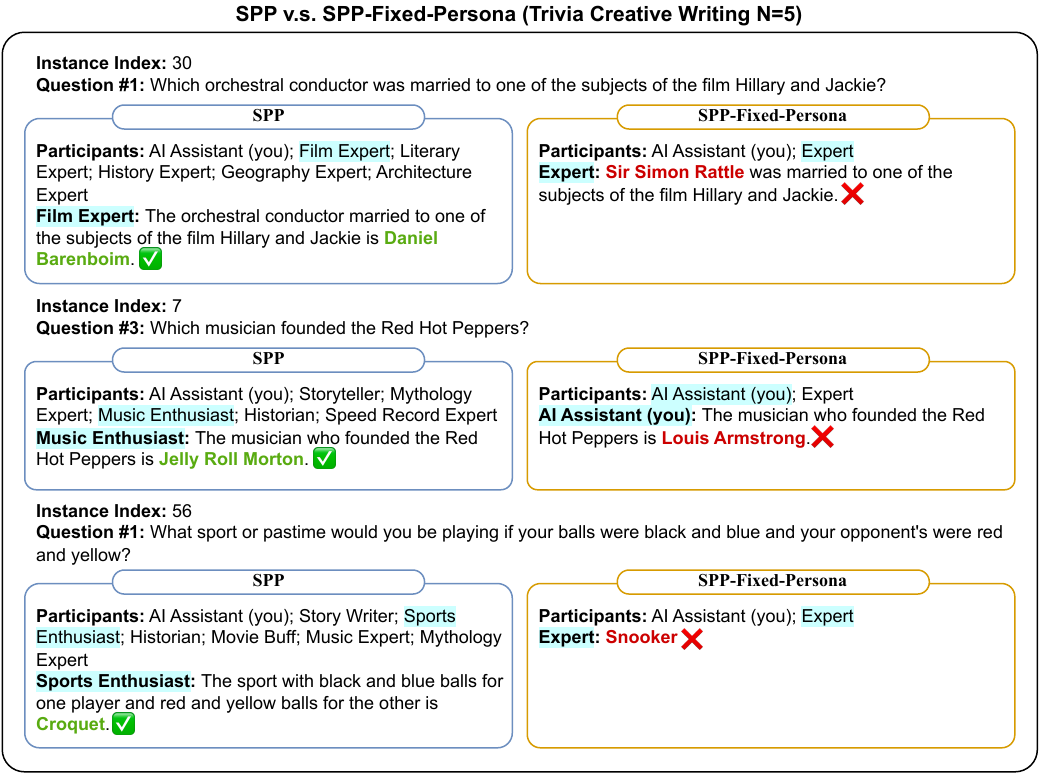}
  \caption{
  SPP vs SPP-Fixed-Persona qualitative examples on \tasktrivia{} (N=5). Each example shows one of the trivia questions in the input instance, the identified participants and the provided answer. We observe that the dynamically identified fine-grained personas, such as "Film Expert", "Music Enthusiast" and "Sports Enthusiast", tend to outperform the fixed general personas, "Expert".
  }
  \label{fig:spp-fixed-qualitative}
\end{figure*}

Figure~\ref{fig:trivia-spp-vs-cot-qualitative} presents examples of the \tasktrivia{} task, illustrating that although CoT can generate plausible plans for task resolution, the final outcomes often contain factual inaccuracies and instances of hallucination. In contrast, \oursabbr{} elicits precise knowledge with fine-grained personas.

Figure~\ref{fig:codenames-spp-vs-cot-qualitative} displays examples of the \taskcodenames{} task, illustrating that \oursabbr{} generates intermediate dialogues that are both \textit{detailed} and \textit{interpretable}, leading to superior performance compared to CoT.

Figure~\ref{fig:spp-fixed-qualitative} shows additional qualitative examples on \ours{} vs \oursproabbr{}.

\section{Early-termination with \oursfixabbr{}}
\label{app:early-termination}


\begin{figure*}[t]
  \centering
  \includegraphics[width=0.95\textwidth]{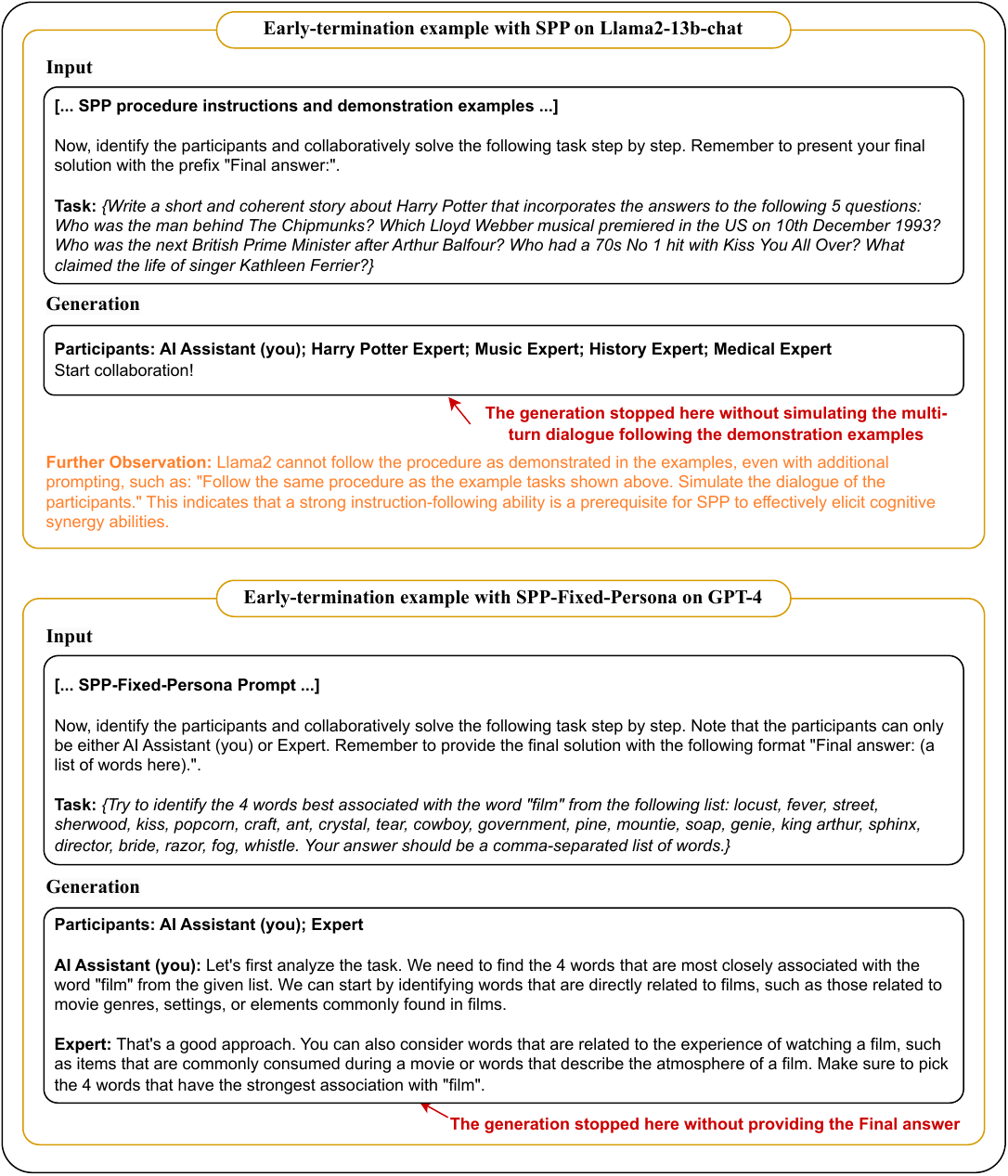}
  \caption{
    Examples of the early-termination problem with \oursabbr{} on Llama2-13b-chat and \oursfixabbr{} on GPT-4.
  }
  \label{fig:early-termination-example}
\end{figure*}
\begin{table*}[t]
    \centering

    \begin{tabular}{lcc}  
      \toprule  
      \textbf{Tasks} & \textbf{added system message} & \textbf{\# early-termination}  \\
      \midrule  
          \multirow{2}{*}{\tasktrivia{} (N=5)}                 &   yes  &  18 / 100 \\
                                                               &   no   &   0 / 100 \\
      \midrule
          \multirow{2}{*}{\tasktrivia{} (N=10)}                &   yes  &  16 / 100 \\
                                                               &   no   &   1 / 100 \\
      \midrule
      \multirow{2}{*}{\taskcodenames{}}                        &   yes  &  37 / 50 \\
                                                               &   no   &   4 / 50 \\
      \midrule
          \multirow{2}{*}{\tasklogic{}}                        &   yes  &  11 / 200 \\
                                                               &   no   &  15 / 200 \\
      \bottomrule  
    \end{tabular}
\caption{Early termination statistics on \textit{\oursfixabbr{}}: Removing the system message, \texttt{"You are an AI assistant that helps people find information."}, can effectively reduce the problem but cannot fully eliminate it.}
  \label{tab:app:early-termination-log}  
\end{table*}  

Figure~\ref{fig:early-termination-example} shows an example of the early-termination problem (defined in \S~\ref{sec:analysis}) where the generation stops before reaching the final solution as if the models is waiting input from an external user.

The problem is particularly severe on certain tasks, e.g., \taskcodenames{}, resulting in unexpectedly low performance as shown in Figure~\ref{fig:analysis_merged_b}.
The problem can be largely alleviated by removing the system message but cannot be entirely eliminated. Table~\ref{tab:app:early-termination-log} shows the statistics of the early-termination problem for each task and method. In contrast, we did not observe early-termination on \oursabbr{}, \oursproabbr{}, Standard, or CoT prompting with GPT-4.

\section{Full Results}
\label{app:full_results}
Full results of the three tasks: \tasktrivia{}, \taskcodenames{} and \tasklogic{} can be found in Tables~\ref{tab:app:trivia_task_full_results}, \ref{tab:app:codenames_task_full_results} and \ref{tab:app:logic_task_full_results}, respectively.
\begin{table*}[t]
    \centering  
    \begin{tabular}{l c c c c}  
      \toprule  
      \multirow{2}{*}{\textbf{Methods}}
      & \multicolumn{4}{c}{\textbf{Scores (N = 5) (\%) }}\\
      & \textbf{w/ system message} & \textbf{w/o system message}  & \textbf{average} & \textbf{max} \\
      \midrule  
          Standard                                       & 75.6 & 73.6 & 74.6 & 75.6 \\
          CoT                                            & 68.8 & 65.6 & 67.1 & 68.8 \\
          \midrule
          Self-Refine [iter=0] & 74.9 & 72.7 & 73.8 & 74.9 \\
          Self-Refine [iter=1] & 75.3 & 72.5 & 73.9 & 75.3 \\
          \midrule
          SPP-Fixed-Persona                              & 66.1 & 79.6 & 72.9 & 79.6 \\
          SPP-Profile                    & 79.8 & 78.3 & 79.1 & 79.8 \\
          \textbf{SPP}                           & \textbf{80.0} & \textbf{79.8} & \textbf{79.9} & \textbf{80.0} \\
      \bottomrule  
    \end{tabular}

    \vspace{0.2cm}
    
    \begin{tabular}{l c c c c }  
      \toprule  
      \multirow{2}{*}{\textbf{Methods}}
      & \multicolumn{4}{c}{\textbf{Scores (N = 10) (\%)}}\\
      & \textbf{w/ system message} & \textbf{w/o system message} & \textbf{average} & \textbf{max} \\
      \midrule  
          Standard                                       & 77.2 & 76.8 & 77.0 & 77.2 \\
          CoT                                            & 71.6 & 65.3 & 68.5 & 71.6 \\
          \midrule
          Self-Refine [iter=0] & 77.1 & 75.4 & 76.3 & 77.1 \\
          Self-Refine [iter=1] & 78.2 & 75.6 & 76.9 & 78.2 \\
          \midrule
          SPP-Fixed-Persona                              & 70.5 & 81.3 & 75.9 & 81.3 \\
          SPP-Profile                    & 82.3 & 83.8 & 83.0 & 83.8 \\
          \textbf{SPP}                           & \textbf{85.2} & \textbf{84.2} & \textbf{84.7} & \textbf{85.2}\\
      \bottomrule  
    \end{tabular}

    \caption{\tasktrivia{} full results, including two inference settings: with system message and without system message. "average" and "max" indicating the mean and max score across the two settings. The system message we use is: \texttt{``You are an AI assistant that helps people find information.''} }
  \label{tab:app:trivia_task_full_results}  
\end{table*}  
\begin{table*}[t]
    \centering  
    \begin{tabular}{l c c c c }  
      \toprule  
      \multirow{2}{*}{\textbf{Methods}}
      & \multicolumn{4}{c}{\textbf{Scores (\%)}}\\
      & \textbf{w/ system message} & \textbf{w/o system message} & \textbf{average} & \textbf{max} \\
      \midrule  
          Standard                                       & 74.5 & \textbf{76.3} & 75.4 & 76.3\\
          CoT                                            & 71.4 & 74.0 & 72.7 & 74.0 \\
          \midrule
          Self-Refine [iter=0] & 77.3 & 73.2 & 75.3 & 77.3 \\
          Self-Refine [iter=1] & 70.1 & 58.8 & 64.4 & 70.1 \\
          \midrule
          SPP-Fixed-Persona                              & 10.1 & 66.0 & 38.1 & 66.0 \\
          SPP-Profile                    & 80.4 & 72.9 & 76.7 & 80.4 \\
          \textbf{SPP }                           & \textbf{82.5} & 75.5 & \textbf{79.0} & \textbf{82.5} \\
      \bottomrule  
    \end{tabular}

  \caption{\taskcodenames{} full results, including two inference settings: with system message and without system message. "average" and "max" indicating the mean and max score across the two settings. The system message we use is: \texttt{``You are an AI assistant that helps people find information.''} }
    
  \label{tab:app:codenames_task_full_results}  
\end{table*}  
\begin{table*}[t]
    \centering  
    
    \begin{tabular}{l c c c c}  
      \toprule  
      \multirow{2}{*}{\textbf{Methods}}
      & \multicolumn{4}{c}{\textbf{Scores (\%)} }\\
      & \textbf{w/ system message} & \textbf{w/o system message} & \textbf{average} & \textbf{max} \\
      \midrule  
          Standard                                       & 56.8 & 58.6 & 57.7 & 58.6 \\
          CoT                                            & \textbf{69.5} & 62.1 & 65.8 & 69.5 \\
          \midrule
          Self-Refine [iter=0] & 62.0 & 55.5 & 58.8 & 62.0 \\
          Self-Refine [iter=1] & 64.5 & 55.5 & 60.0 & 64.5 \\
          \midrule
          SPP-Fixed-Persona                              & 63.3 & 65.3 & 64.3 & 65.3 \\
          SPP-Profile                    & 65.7 & 64.0 & 64.8 & 65.7\\
          \textbf{SPP}                           & 66.3 & \textbf{70.4} & \textbf{68.3} & \textbf{70.4}\\
      \bottomrule  
    \end{tabular}
    \caption{\tasklogic{} full results, including two inference settings: with system message and without system message. "average" and "max" indicating the mean and max score across the two settings. The system message we use is: \texttt{``You are an AI assistant that helps people find information.''} }
  \label{tab:app:logic_task_full_results}  
\end{table*}

\section{Usage of AI assistants in writing}
We used ChatGPT and GPT-4 solely for checking and correcting grammars.

\end{document}